\documentclass[]{article}
\usepackage{a4wide}

\usepackage{times}  
\usepackage{helvet}  
\usepackage{courier}  
\usepackage[hyphens]{url} 
\usepackage{graphicx} 
\usepackage{caption} 

\usepackage{subcaption}

\usepackage{soul}

\usepackage{algorithm}
\usepackage{algorithmic}

\usepackage{amsthm}
\usepackage{amssymb}
\usepackage{todonotes}

\usepackage{tikz} 

\usepackage{enumitem}

\usepackage{float}

\newtheorem{example}{Example}
\newtheorem{definition}{Definition}
\newtheorem{proposition}{Proposition}

\newcommand{\ep}{{\sf Ep}}
\newcommand{\ip}{{\sf Ip}}
\newcommand{\ec}{{\sf Ec}}
\newcommand{\ic}{{\sf Ic}}
\newcommand{\cn}{{\sf Cn}}
\newcommand{\ex}{{\sf Ex}}
\newcommand{\pe}{{\sf Pe}}

\newcommand{\epset}{W^{\sf p}}
\newcommand{\ipset}{D^{\sf p}}
\newcommand{\ecset}{W^{\sf c}}
\newcommand{\icset}{D^{\sf c}}

\newcommand{\defaultargument}{\langle\epset,\ipset,\ecset,\icset\rangle}

\newcommand{\support}{{\sf S}}
\newcommand{\consequence}{{\sf C}}

\newcommand{\justification}{{\sf Jp}}

\newcommand{\nodes}{{\sf Nodes}}

\newcommand{\args}{{\sf Args}}

\newcommand{\mynull}{\mbox{\em Null}}

\newcommand{\myblue}[1] {\textcolor{blue}{#1}}
\newcommand{\myred}[1] {\textcolor{red}{#1}}

\newcommand{\qw}[1] {{\tt {#1}}}

\title{\noindent\rule{\textwidth}{4pt}\\Understanding Enthymemes in Argument Maps:\\
Bridging Argument Mining and Logic-based Argumentation\\\noindent\rule{\textwidth}{1pt}}

\author{
  Jonathan Ben-Naim$^{1}$ \and
  Victor David$^{2}$ \and
  Anthony Hunter$^{3}$ \and  
$^1$Universit\'{e} Paul Sabatier, CNRS, IRIT - Toulouse, France \\
$^2$Universit\'{e} Côte d'Azur, Inria, I3S - Sophia Antipolis, France \\
$^3$University College London - London, UK 
}

\date{}

\begin{document}

\maketitle

\begin{abstract}
Argument mining is natural language processing technology aimed at identifying arguments in text. Furthermore, the approach is being developed to identify the premises and claims of those arguments, 
and to identify the relationships between arguments including support and attack relationships.
In this paper, we assume that an argument map contains the premises and claims of arguments, and support and attack relationships between them, that have been identified by argument mining.
So from a piece of text, we assume an argument map is obtained automatically by natural language processing.
However, to understand and to automatically analyse that argument map, 
it would be desirable to instantiate that argument map with logical arguments.
Once we have the logical representation of the arguments in an argument map, 
we can use automated reasoning to analyze the argumentation (e.g. check consistency of premises, check validity of claims, and check the labelling on each arc corresponds with the logical arguments). 
We address this need by using classical logic for representing the explicit information in the text, 
and using default logic for representing the implicit information in the text.
In order to investigate our proposal, 
we consider some specific options for instantiation. 
\end{abstract}
\section{Introduction}

\begin{figure}
\begin{center}
\begin{tikzpicture}
[->,>=latex,thick, scale=0.6,
txtarg/.style={draw,text centered, text width=35mm,
shape=rectangle, rounded corners=2pt,
fill=gray!20,font=\footnotesize},
]
\node[txtarg] (a0) [] at (0,5) {n0: Cars should be banned from cities.};
\node[txtarg] (a1) [] at (0,2.5) {n1: Cars are polluting, and so bad for the health.};
\node[txtarg] (a2) [] at (8,5) {n2: Cars are vital for people to move around cities.};
\node[txtarg] (a3) [] at (0,0) {n3: Internal combustion engines pollute.};
\node[txtarg] (a4) [] at (8,2.5) {n4: Because of legislation, soon all new cars will be electric.};
\path	(a1) edge node[left] {$+$} (a0);
\path	(a2) edge node[above] {$-$} (a0);
\path	(a3) edge node[left] {$+$} (a1);
\path	(a4) edge node[above] {$-$} (a1);
\end{tikzpicture}
\end{center}
\caption{\label{fig:intrograph}Example of an argument graph where 
each node in $\{n0,n1,n2,n3,n4\}$ represents an argument, 
and the text exposition is given after the colon.
The $+$ (resp. $-$) label on an arc denotes a support (resp. attack) relationship.
According to the text, $n0$ appears to be a claim without premises, though $n1$ appears to provide support for what might be implicit premises of $n0$. Also $n1$ appears to have premises for its own claim. So what are the implicit premises of $n0$ and how do they entail the claim of $n0$? And what does the claim of $n1$ entail for supporting the implicit premise of $n0$? We can also ask about how the claim of $n2$ attacks $n0$. Is $n2$ attacking the claim of $a0$ or is it attacking the implicit premises of $n0$? We can ask the same kind of questions for the remaining arguments and arcs here. 
}
\end{figure}
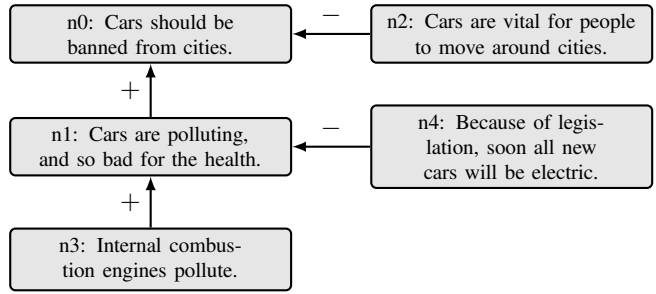

\begin{figure}
\begin{center}
\begin{tikzpicture}
[->,>=latex,thick, scale=0.6,
txtarg/.style={draw,text centered, text width=35mm,
shape=rectangle, rounded corners=2pt,
font=\footnotesize},
]
\node[txtarg] (a0) [] at (0,5) {n0: (P) \myred{Null}; (C) \myblue{Cars should be banned from cities}};
\node[txtarg] (a1) [] at (0,2.5) {n1: (P) \myred{Cars are polluting}; (C) \myblue{Cars are bad for the health}};
\node[txtarg] (a2) [] at (8,5) {n2: (P) \myred{Null}; (C) \myblue{Cars are vital for people to move around cities}};
\node[txtarg] (a3) [] at (0,0) {n3: (P) \myred{Null}; (C) \myblue{Internal combustion engines pollute}};
\node[txtarg] (a4) [] at (8,2.5) {n4: (P) \myred{Legislation}; (C) \myblue{Soon all new cars will be electric}};
\path	(a1) edge node[left] {$+$} (a0);
\path	(a2) edge node[above] {$-$} (a0);
\path	(a3) edge node[left] {$+$} (a1);
\path	(a4) edge node[above] {$-$} (a1);
\end{tikzpicture}
\end{center}
\caption{\label{fig:intrograph2}Continuing Figure \ref{fig:intrograph}, an argument map where 
each node in $\{n0,n1,n2,n3,n4\}$ is an argument.
After the colon, the text for the premise follows (P) and text for the claim follows (C).
The $+$ (resp. $-$) on an arc denotes a support (resp. attack) relationship.
}
\end{figure}
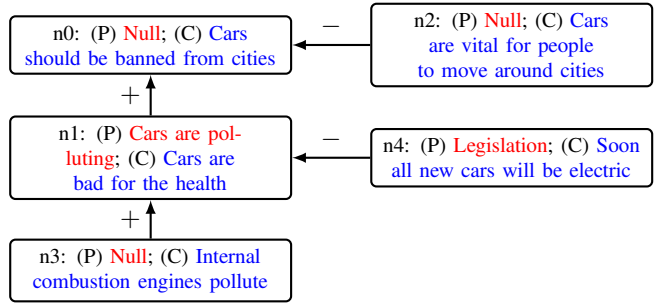

Argument mining aims to identify the parts of text that represent arguments (premises and/or claims), and the support or attack relationships between those arguments \cite{Lawrence2019}. Using a range of natural language processing (NLP) technology, with increasing emphasis on large language models, they can be trained to identify arguments and relationships between them with increasing accuracy. The resulting information can then be represented by an argument graph or argument map, where each node contains the text representing the premises and claim of the argument, and each arc denotes a relationship (such as positive/support, or negative/attack) between arguments, as illustrated in Figures \ref{fig:intrograph} and \ref{fig:intrograph2}. In this paper, we assume the difference between an argument graph and argument map is the latter distinguishes between the premises and the claim of each argument. 

Whilst argument graphs and maps 
provide a useful visualizable summary of argumentation in text, there is then a lack of automated methods for analysing or reasoning with the information in the argument map. To address this shortcoming, it would be desirable to translate the text into logical arguments, and then use automated reasoning to analyze the arguments
(e.g. check consistency of premises, check validity of claims, and check the labelling on each arc corresponds with the logical arguments assigned to the abstract arguments in the map). 
Unfortunately, there is a lack of a formal representational framework for bridging argument maps and the underlying logical arguments.
So this gives us the first problem that we tackle in this paper: 

\begin{quote}
({\bf Problem 1}) {\em How can we represent the explicit information in an argument map, and then represent that explicit information in a logical rendition of the arguments?}
\end{quote}

There are a number of frameworks for modelling argumentation in logic (e.g. \cite{BH05,Gorogiannis2011,Modgil2014,Toni2014,amgoud2018measuring,cayrol2020logical}). 
They incorporate a formal representation of individual arguments, where the premises imply the claim, and techniques for comparing conflicting arguments \cite{Atkinson2017}. 
However, real arguments (i.e. arguments presented by humans) usually do not have enough explicitly presented premises for the  entailment of the claim. 
This is because there is some common or commonsense knowledge\footnote{Commonsense knowledge is knowledge normally known by everyone, i.e. universally known (as opposed to local knowledge). For instance, it is commonsense knowledge that if drop an egg on the floor, it will probably break, or if you tell your friends a funny joke, they will probably laugh. For a review of commonsense knowledge representation and reasoning see \cite{Davis2017}.
In contrast, common knowledge is knowledge that is known by a subset of people?
For example, if two old friends are talking and one of them refers a past event that they had experienced together, without explicitly recalling it, they may both have implicit but common knowledge of the event which would be a context known only to relatively few people (non-universal knowledge).}
that can be assumed by a proponent of an argument and the recipient of it \cite{Walton2001}.

Whilst human agents constantly need to understand enthymemes, whether in everyday or professional life,  
there is a lack of adequate algorithmic methods for supporting or automating this process. 
The coding/decoding can be modelled as abduction \cite{Hunter2007,Black2012,Hosseini2014},
and there has been some consideration of how this can be undertaken within a dialogue \cite{Black2008,Dupin2011,Dupin2011b,Xydis2020,Panisson2022,Leiva2023}.
However, these proposals do not provide a systematic framework for translating argument maps into logic, and concomitantly, addressing the problem of identifying the missing premises and/or claim, and discerning the relationships between them. 
We illustrate this problem with example of an argument graph in Figure \ref{fig:intrograph},
and in the argument map in  Figure \ref{fig:intrograph2} where we flag the premises and claims that are completely implicit with the ``Null" value. This brings us to the second problem we tackle in this paper:

\begin{quote}
({\bf Problem 2}) {\em How can we represent the implicit information pertaining to the arguments in an argument map within a logical rendition of the arguments?}
\end{quote}

To address these two questions in this paper, we start from the assumption that each premise, and each claim, is a string of text, such as a phrase, a sentence, or a paragraph. 
If the premise or claim is implicit, then we use the Null value. 
We then assume a function that can translate each piece of text into a formula of classical logic, where the formula is an atom, a propositional formula, or a first-order formula, depending on the nature and granularity of translation. We use classical logic because it has a clear syntax and semantics for representing and reasoning with information. 
Then, for each argument we use default logic (which we introduce in the next paragraph) for connecting each claim with its premise, and each argument with the arguments it supports or attacks.

Default logic was developed as a formalism for commonsense reasoning. It extends classical logic with default rules \cite{Reiter1980}. Default rules are a generalization of natural deduction rules. They provide a clear representation of how default inferences can be obtained, and of how this reasoning can be attacked. Whilst there are previous proposals for using default logic in logic-based argumentation (e.g. \cite{Prakken1993,Santos2008,Hunter2018}), none of these proposals provide a comprehensive coverage of the diversity of attack and support relations that can be captured, and moreover, none of these proposals consider how implicit information as arising in enthymemes can be handled, nor how argument maps can be instantiated by logic-based arguments.

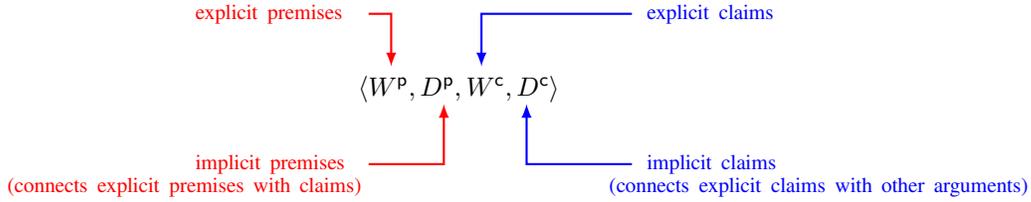
\begin{figure}[h]
\centering
\begin{tikzpicture}
[->,>=latex,thick, scale=1]
]
\node[] (a) [] at (2,1) {$\langle\epset,\ipset,\ecset,\icset\rangle$};
\node[text width=30mm] (b) [] at (0,2) {\myred{\footnotesize explicit premises}};
\node[text width=30mm] (c) [] at (0,0) {\myred{\footnotesize implicit premises}};
\node[text width=30mm] (d) [] at (6,2) {\myblue{\footnotesize explicit claims}};
\node[text width=30mm] (e) [] at (6,0) {\myblue{\footnotesize implicit claims}};
\node[text width=80mm] (e) [] at (0,-0.3) {\myred{\footnotesize (connects explicit premises with claims)}};
\node[text width=60mm] (e) [] at (7,-0.3) {\myblue{\footnotesize (connects explicit claims with other arguments)}};
\draw[red] (0.8,2) |- (1.1,2) -| (1.1,1.3);
\draw[red] (0.8,0) |- (1.8,0) -| (1.8,0.8);
\draw[blue] (4.3,2) |- (2.3,2) -| (2.3,1.3);
\draw[blue] (4.3,0) |- (2.9,0) -| (2.9,0.8);
\end{tikzpicture}
\caption{Structure of a default argument}
\end{figure}

We introduce the notion of a default argument as a tuple $\defaultargument$ where $\epset$ is a set of classical formulae that represents the explicit premises of the argument, $\ipset$ is a set of default rules that denote the implicit premises of the argument, $\ecset$ is a set of classical formulae that represent the explicit claims of the argument, and $\icset$ is a set of default rules that denote the implicit claim of the argument. 

For each node in the argument map, we identify an appropriate logical argument to instantiate the node. This results in an instantiated argument map which can be analysed to check the validity of arguments, and of the labels on the arcs. Furthermore, different instantiations for the same argument map can be compared (e.g. to determine whether one is finer-grained than another), and different maps can be compared in terms of their instantiated version (e.g. to determine whether one is equivalent to another in some sense).  This brings us to the third problem we tackle in this paper: 

\begin{quote}
({\bf Problem 3}) {\em How do we analyze and compare argument maps that are instantiated with default arguments?}
\end{quote}

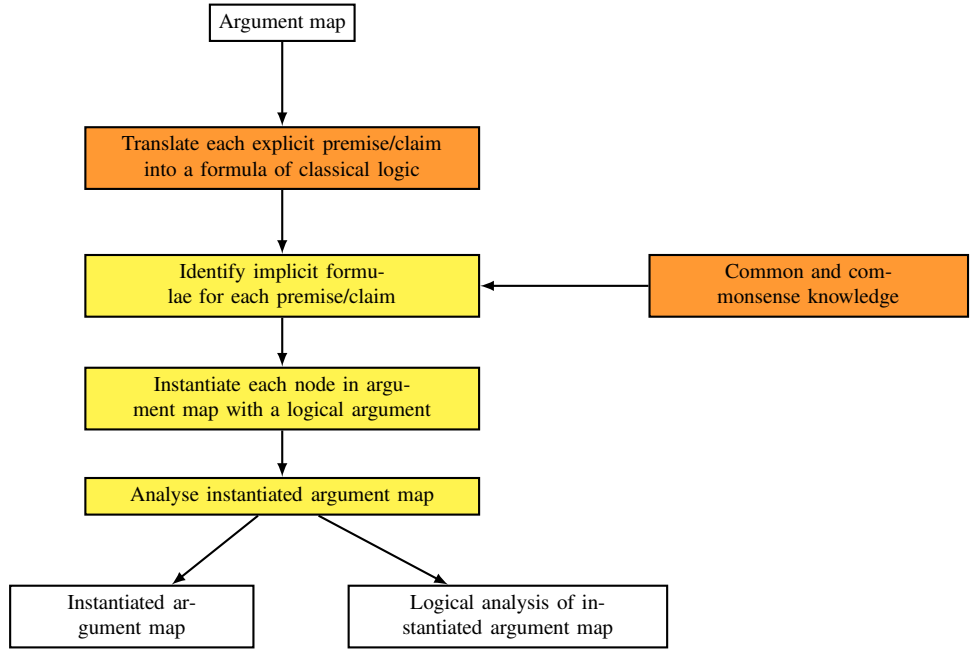
\begin{figure}[t]
\begin{center}
\begin{tikzpicture}
[->,>=latex,thick, 
mynode/.style={draw,text centered, text width=50mm,
shape=rectangle,fill=yellow!80,
font=\footnotesize},
mynode2/.style={draw,text centered, 
shape=rectangle,font=\footnotesize}
]
\node[mynode2] (a) [] at (0,7.5) {Argument map};
\node[mynode,fill=orange!80] (b) [] at (0,5.7) {Translate each explicit premise/claim into a formula of classical logic};
\node[mynode] (c) [] at (0,4) {Identify implicit formulae for each premise/claim};
\node[mynode] (d) [] at (0,2.5) {Instantiate each node in argument map with a logical argument};
\node[mynode] (e) [] at (0,1.2) {Analyse instantiated argument map};
\node[mynode2,text width=30mm] (f1) [] at (-2,-0.4) {Instantiated argument map};
\node[mynode2,text width=40mm] (f2) [] at (3,-0.4) {Logical analysis of instantiated argument map};
\node[mynode2,fill=orange!80,text width=40mm] (g) [] at (7,4) {Common and commonsense knowledge};
\path	(a) edge node[] {} (b);
\path	(b) edge node[] {} (c);
\path	(c) edge node[] {} (d);
\path	(d) edge node[] {} (e);
\path	(e) edge node[] {} (f1);
\path	(e) edge node[] {} (f2);
\path	(g) edge node[] {} (c);
\end{tikzpicture}
\end{center}
\caption{\label{fig:overview}Overview of our pipeline for understanding enthymemes. The input is the white box at the top, and the outputs are the the white boxes at the bottom. The focus of this paper is on the knowledge representation and reasoning aspects of the yellow boxes (i.e. how do we represent the explicit and implicit aspects of a logical argument, how do we represent the instantiation of an argument map with logical arguments, and what kinds of logical analyse can we undertake with the instantiation of an argument map). We leave the underlying mechanisms for these yellow boxes to future work (i.e. algorithms for identifying implicit formulae for each premise and claim, algorithms for instantiating each node in argument map with a logical argument, and algorithms for automated analysis of instantiated argument maps). We also leave the orange boxes to future work (i.e. the automated natural language understanding of premises and claims of natural language arguments, the acquisition of commonsense knowledge, and knowledge representation and reasoning with commonsense knowledge).}
\end{figure}

In the rest of this paper, we review default logic, and then use it to define the notion of a default argument, and to define different notions of support and attack between arguments. We then define how we bridge argument maps and logical argumentation by translating the sentences in argument maps into classical formulae, and then instantiating the argument maps with logical arguments. 
Once we have presented our novel framework, we compare it with the related literature. We conclude with a discussion of our proposal and of future work. We provide a summary overview of our pipeline in Figure \ref{fig:overview}.

\section{Translation of explicit premise and claim into logic}
\label{section:translation}

We assume the usual propositional and predicate (first-order) languages for classical logic.
Let ${\cal A}$ be the set of ground atoms (i.e. propositional atoms and positive ground literals).
Let ${\cal P}$ be the set of propositional formulae
composed from atoms ${\cal A}$  and the logical connectives $\wedge$, $\vee$, $\neg$.
Let ${\cal F}$ be the set of first-order formulae
composed from a set of function symbols, predicate symbols, the logical connectives $\wedge$, $\vee$, $\neg$,
and the quantifiers $\forall$ and $\exists$.
Since we are often indifferent as to whether we use the propositional or first-order language, we use $\cal L$ to denote a language that can be either a propositional or first-order language. In the examples, we will use teletype font for the logical formulae.

\subsection{Argument maps}

We assume an argument map is obtained as output from argument mining of a text (e.g. a discussion document).
Each node denotes an argument (often with some/all of the premises and/or claim being implicit).
Often the text in a node appear as a statement rather than a complete argument since either the premises or claim are implicit.
We represent an argument map as a tuple as follows.

\begin{definition}
Let ${\cal T}$ be a set of text strings.
An {\bf argument map} is a tuple $(N,P,C,L)$
where
$N$ is a set of nodes;
$P: N \rightarrow {\cal T} \cup \{\mynull \}$ is a text labelling function for premises;
$C: N \rightarrow {\cal T} \cup \{\mynull \}$ is a text labelling function for claims;
And $L: N \times N \rightarrow \{+,-,\Box \}$ 
is an arc labelling function, 
where $+$ represents an attack relationship, 
$-$ represents a support relationship,
and $\Box$ represent no relationship holds from the first node to the second node.  
\end{definition}

\begin{figure}
\begin{center}
\begin{tabular}{p{2mm}p{55mm}p{65mm}}
\hline
N & P & C \\
\hline
n0 & Null & Cars should be banned from cities.\\
n1 & Cars are polluting & Cars are bad for the health.\\
n2 & Null  & Cars are vital for people to move around cities.\\
n3 & Null & Internal combustion engines pollute.  \\
n4 & Legislation & Soon all new cars will be electric.\\
\hline
\end{tabular}
\end{center}
\caption{An argument map $(N,P,C,L)$ 
where the graph $(N,A)$ and the labelling function $L$ is given in  Figure \ref{fig:intrograph2},
and the $P$ and $C$ functions for the arguments are in the table.}
\end{figure}

Each premise text and each claim text is a phrase, or sentence, or paragraph of text. 
It communicates the explicit information required for the premise or claim of each argument. 
So the starting point for a formalization of argumentation should be the representation of these strings of text. 
There is a long history of graphical representation of arguments and counterarguments where text associated with each argument and counterargument. An argument map is an example of such a representation and we can consider the output of argument mining as being an argument map. 

\subsection{Logical translation}

Given an argument map, we want to represent each of the premise and the claim by a set of formulae of classical logic.
The following definition of translation function is a key step in bridging the output of argument mining (i.e. the text-based annotation of argument maps) and a logic-based representation of the argument map.

\begin{definition}
Let ${\cal T}$ be a set of text strings.
A {\bf logical translation} 
is a function $T:{\cal T}\cup \{\mynull\} \rightarrow \wp({\cal L})$
s.t.\footnote{$\wp$ stands for power-set.} $T(\mynull) = \emptyset$. 
\end{definition}

If there is no text string (i.e. Null), then the translation is $\emptyset$ (i.e. tautology) which represents no useful information.

\begin{example}
\label{ex:translation:atomic}
Containing Example \ref{fig:intrograph2},
the following logical translation $T$ is listed for the text string for each premise and claim. 
In this example, each text string is assigned an atom in the logical language.
\[
\begin{array}{lll}
n0 & T(\mynull) = \emptyset & T(\mbox{\em Cars should be banned from cities}) = \{\mbox{\tt s1}\}\\
n1 & T(\mbox{\em Cars are polluting}) = \{\mbox{\tt s3}\} & T(\mbox{\em Cars are bad for the health}) = \{\mbox{\tt s0}\}\\
n2 &T(\mynull) = \emptyset  & T(\mbox{\em Cars are vital for people to move around cities}) =\{\mbox{\tt s2}\}\\
n3 &T(\mynull) = \emptyset & T(\mbox{\em Internal combustion engines pollute}) = \{\mbox{\tt s4} \} \\
n4 &T(\mbox{\em Legislation}) = \{\mbox{\tt s5}\}& T(\mbox{\em Soon all new cars will be electric}) =\{\mbox{\tt s6}\}\\
\end{array}
\]
\end{example}

The following are some types of translation, where the first is the most abstract and the simplest to implement, and the third is the least abstract of the three, and it is the most difficult to implement.

\begin{description}

\item {\bf Atomic translation.} The logical translation is a set of propositional atoms. So each text string is assigned an atom (as illustrated in Example \ref{ex:translation:atomic}). This allows for different ways that a claim is expressed in text being assigned to the same atom (as illustrated in Example \ref{ex:translation:atomic2}). 
This can be implemented by a text string classifier where each string is assigned an atom that denotes its class. Assignments can group similar text items to the same atom as in intent classification \cite{Gretz2022} or key point analysis \cite{BarHaim2021}. Also, in an argumentative chatbot system, sentence embeddings have been used to classify a user input as to being one of around 1500 arguments in an argument graph in order to identify a counterargument to be given by the chatbot \cite{Chalaguine2020}.

\item {\bf Propositional translation.} The logical translation is the set of propositional formulae that can be formed from a set of propositional atoms (as illustrated in Example \ref{ex:translation:propositional}). A translation function can be implemented using natural language processing that can identify the underlying Boolean connectives in a sentence (e.g.\cite{Wyner2016,Lu2022}).

\item {\bf First-order translation.} The logical translation is the set of first-order formulae that can be formed from a set of predicate and function symbols (as illustrated in Example \ref{ex:translation:firstorder}). A translation function could then be implemented using natural language processing such as an AMR parser \cite{Damonte2017,Drozdov2022}, neural network \cite{Singh2020}, or reinforcement learning \cite{Lu2022}. 

\end{description}

\begin{example}
\label{ex:translation:atomic2}
Consider the text for two arguments ``we should stop quantitative easing because government printing of money can cause prices to rise"
and `we should stop quantitative easing because quantitative easing can cause inflation". 
Assume that by using argument mining applied to the first argument, the text for the premise is ``government printing of money can cause prices to rise" 
and applied to the second argument, the text for the premise is ``quantitative easing can cause inflation". 
These premises may be regarded as making the same point. 
This can be implemented by a text string classifier where each string is assigned an atom that denotes its class.
\[
\begin{array}{l}
T(\mbox{``government printing of money can cause prices to rise"}) = {\tt a_{13}}\\
T(\mbox{``quantitative easing can cause inflation"}) = {\tt a_{13}}\\
\end{array}
\]
\end{example}

\begin{example}
\label{ex:translation:propositional}
Consider the text for an argument  ``quantitative easing causes inflation, which is highly undesirable, and so we should stop quantitative easing". 
Assume that by using argument mining applied to this argument, the text for the premise is ``quantitative easing causes inflation, which is highly undesirable",
and the text for the claim is ``we should stop quantitative easing". 
The following is a propositional translation of the premise and claim.
\[
\begin{array}{l}
T(\mbox{``quantitative easing causes inflation, which is highly undesirable"}) = \\
\hspace{4cm} \tt (quantitative\_easing \rightarrow inflation) \wedge \neg inflation\\
\\
T(\mbox{``we should stop quantitative easing"}) = \neg {\tt quantitative\_easing}\\
\end{array}
\]
\end{example}

\begin{example}
\label{ex:translation:firstorder}
Consider the text for an argument  ``we should stop quantitative easing because whenever a country uses long-term quantitative easing, there is high inflation". 
Assume that by using argument mining applied to this argument, the text for the premise is ``whenever a country uses long-term quantitative easing, there is high inflation". 
The following is a first-order translation of the premise.
\[
\begin{array}{l}
T(\mbox{``whenever a country uses long-term quantitative easing, there is high inflation"}) = \\
\hspace{2cm} \tt \forall x \; country(x) \wedge quantitative\_easing(x,long\_term) \rightarrow inflation(x,high)
\end{array}
\]
\end{example}

The size of the codomain of the logical translation relative to the domain determines the granularity of the translation. For example, if the domain is large, but we only have relatively few formulae in the codomain, then we have a coarse-grained translation.

\section{Review of default logic}

We use $\alpha,\beta,\gamma,\delta,\phi,\psi,\ldots$ for arbitrary formulae
and $\Delta, \Gamma, \ldots$ for arbitrary sets of classical formulae.
We let $\vdash$ denote the classical consequence relation, 
and write $\Delta \vdash\bot$ to denote that $\Delta$ is inconsistent.
${\sf Atoms}(\Delta)$ gives the atoms appearing in the formulae in $\Delta$. 
Let $\cn$ be the consequence closure function (i.e.  $\cn(\Delta) = \{ \phi \mid \Delta \vdash \phi \}$). 
For $\phi,\psi\in {\cal L}$, $\phi\equiv\psi$ denotes that $\phi$ and $\psi$ are equivalent 
(i.e. $\{\phi\}\vdash\psi$ and $\{\psi\}\vdash\phi$). 
For $\Delta,\Gamma\subseteq {\cal L}$, $\Delta\equiv\Gamma$
 denotes that $\Delta$ and $\Gamma$ are equivalent (i.e. $\cn(\Delta) = \cn(\Gamma)$).

As a basis of representing and reasoning with default knowledge,
default logic, proposed by Reiter \cite{Reiter1980}, is one of the best known and most widely studied
formalisations of default reasoning. Furthermore, it offers a very expressive and
lucid language.  

In default logic, knowledge is represented as a set of propositional or first-order formulae and a
set of default rules for representing default information. 
A {\bf default rule} is of the following form, where $\alpha$,
$\beta$ and $\gamma$ are classical formulae.
\[
\frac{\alpha : \beta}{\gamma}
\]
For
this, $\alpha$ is called the {\bf pre-condition}, $\beta$ is called the
{\bf justification}, and $\gamma$ is called the {\bf consequent}, of the default rule.

A {\bf default theory} is a pair $(D,W)$ where $D$ is a set of default rules and $W$ is a set of classical formulae.
Default logic extends classical logic. Hence, all classical inferences from the classical information in a default theory are derivable (if there is an extension as defined below). The default theory then augments these classical inferences by default inferences derivable using the default rules: If $\alpha$ is
inferred, and $\neg\beta$ cannot be inferred, then infer $\gamma$. 
The following is a definition for when a default theory has an extension.

\begin{definition}
\label{defi:extension}
Let $(D,W)$ be a default
theory, where D is a set of default rules and W is a set of classical
formulae. 
The operator $\pe$ indicates what
conclusions are to be associated with a given set $E$ of formulae,
where $E$ is some set of classical formulae. 
For this,
$\pe(E)$ is the smallest set of classical formulae such that the
following three conditions are satisfied. 

\begin{enumerate}

\item $W \subseteq \pe(E)$,

\item $\pe(E) = \cn(\pe(E))$,

\item For each default in $D$, where $\alpha$ is the pre-condition, 
$\beta$ is the justification, 
and $\gamma$ is the consequent, 
the following holds: if $\alpha \in \pe(E)$, 
and $\neg \beta \not \in E$,   
then $\gamma \in \pe(E)$.

\end{enumerate}

We refer to $E$ as the {\bf satisfaction set}, 
and $\pe(E)$ the {\bf potential extension}.
Furthermore, $E$ is an {\bf extension} of $(D,W)$ iff $E=\pe(E)$.
\end{definition}

We explain the above definition as follows:
if $E$ is an extension, then the first condition ensures that the set of classical formulae $W$ is also in the extension, the second condition ensures the
extension is closed under classical consequence, and the third
condition ensures that for each default rule, if the pre-condition is
in the extension, and the justification is consistent with the
extension, then the consequent is in the extension. 
Minimality ensures that the extension is grounded (i.e. it avoids one or more default rules being self-supporting).

\begin{example}
\label{ex:tweety}
Let $D$ be the following set of defaults. Note, each default in this example is given as a schema where the variables in the default rule are grounded before use to give a set of propositional default rules.
\[
\frac{\tt bird(X) : \neg penguin(X) \wedge fly(X)}{\tt fly(X)}
\hspace{1cm}
\frac{\tt penguin(X) : bird(X)}{\tt bird(X)}
\hspace{1cm}
\frac{\tt penguin(X) : \neg fly(X)}{\tt \neg fly(X)}
\]
For $(D,W)$, where  $W$ is $\{\tt bird(Tweety) \}$, we obtain one extension 
\[
\cn(\{\tt bird(Tweety), fly(Tweety) \})
\]
For $(D,W)$, where $W$ is $\{\tt penguin(Tweety) \}$, we obtain one extension 
\[
\cn(\{\tt penguin(Tweety), bird(Tweety), \neg fly(Tweety)\})
\]
\end{example}

\begin{example}
\label{ex:tweety2}
Let $D$ be the following set of defaults. 
\[
\frac{\tt bird(X) : fly(X)}{\tt fly(X)}
\hspace{1cm}
\frac{\tt penguin(X) : \neg fly(X)}{\tt \neg fly(X)}
\]
For $(D,W)$, where  $W$ is $\{\tt bird(Tweety), penguin(Tweety) \}$, we obtain two extensions 
\[
\cn(\{\tt bird(Tweety), fly(Tweety) \})
\hspace{1cm}
\cn(\{\tt penguin(Tweety), \neg fly(Tweety)\})
\]
\end{example}

The notion of an extension in default logic is different to that of the notion of extension in abstract argumentation. In the former, an extension is a set of formulae obtained from a default theory, whereas in the latter, an extensions is an acceptable set of arguments from an argument graph. However, as shown by Bondarenko {\em et al.} \cite{Bondarenko1997}, default logic can be modelled in assumption-based argumentation that is in turn based on abstract argumentation. But despite this underlying relationship between default logic and argumentation, there remains the need to address the questions raised in the introduction of this paper concerning the need for mechanisms for translation of argument maps into structured argumentation, and understand how implicit knowledge can be obtained and represented in order to understand enthymemes. This in turn calls for a more comprehensive study of how different kinds of argument, including enthymemes, and attack and suppport relationships between them, can be represented in a formalism based on default logic. We will explore these issues in the rest of this paper.

\subsection{Types of default theory}

There are various special cases of default rules that will be useful for our purposes. These include the following.

\begin{itemize}

\item {\bf Precondition-free default rule.}
This is a default rule of the form $\top:\beta/\gamma$ and so the consequent is obtained as long as the justification is satisfiable.
For example, we might use the following default rule to capture the reasoning that anything that is unbroken is usable.
\[
\tt \top: \neg broken(X)/useable(X)
\]

\item {\bf Justification-free default rule.} 
This is a default rule of the form $\alpha:\top/\gamma$ and so the consequent is obtained as long as the precondition holds, and so there is no block by the justification. This would mean that we assume that there are no exceptions.
For example, we might use the following default rule to capture the reasoning that anything that is divisible by two is even.
\[
\tt divisablebytwo(X):\top/even(X)
\]

\item {\bf Normal default rule.} 
This is a default rule of the form $\alpha:\beta/\beta$ and so the consequent is obtained when the precondition holds, and the consequent is satisfiable.
For example, we might use the following default rule to capture the reasoning that if it is consistent to believe that someone has no brother, then we infer they have no brother.
\[
\tt  person(X) :\neg hasBrother(X)/\neg hasBrother(X)
\]

\item {\bf Semi-normal default rule.} 
This a default rule of the form $\alpha:\beta\wedge\gamma/\gamma$ and so the consequent is obtained when the precondition holds, and the justification which includes the consequent is satisfiable.
For example, we might use the following default rule to capture that a bird flies if it is consistent to believe that it flies and that it is not a penguin (as we used in Example \ref{ex:tweety}).
\[
\frac{\tt bird(X) : \neg penguin(X) \wedge fly(X)}{\tt fly(X)}
\]
\end{itemize}

For more coverage of default logic, see \cite{Besnard1989,Brewka1991,Brewka1997}.
There are algorithms for automated reasoning with default logic \cite{Besnard1983,Risch1994},
and scalable implementations of default logic \cite{Tammet2022}.
Also, a default theory can be translated into an answer set program (ASP) and an ASP solver used to automate the reasoning \cite{Chen2010}.

\subsection{Singular default theories}

In the next section, we use default logic in our definition for default arguments.
For this, the only restriction is that the default rules in the premises give a unique extension, and for this we introduce the following subsidiary definition.

\begin{definition}
\label{def:singular}
A default theory $(D,W)$ is {\bf singular} 
iff there is a unique extension of $(D,W)$. 
When a default theory $(D,W)$ is singular,
let $\ex(D,W)$ denote the extension. 
\end{definition}

\begin{example}
The default theory $\tt (\{  (a:b/b), (b\vee c: d\wedge f/e)   \}, \{a\})$ is singular,
and so there is a unique extension $\ex(\tt \{  (a:b/b), (b\vee c: d\wedge f/e) \},\{a\}) = \cn(\{a,b,e\})$. 
In contrast, the default theory $\tt (\{a\},\{  (a:b/b), (a: \neg b/\neg b)   \})$ is not singular,
as there are two extensions $\cn(\{\tt a,b\})$ and $\cn(\{\tt a,\neg b\})$. 
\end{example}

The default theories $(\emptyset,W)$ and $(D,\{\bot\})$, for any $D$, are singular. 
For the former, $\ex(\emptyset,W) = \cn(W)$,
and for the latter, $\ex(D,\{\bot\}) = \cn(\{\bot\})$. 
Also, if $(D,W)$ is singular, then for all $D'\subseteq D$, then $(D',W)$ is singular.

\begin{proposition}
(Proposition 6.2.23 in \cite{Besnard1989}) 
For a default theory $(D,W)$, 
$(D,W)$ is singular 
if $W$ is consistent with $\{\beta\wedge\gamma \mid \alpha : \beta / \gamma \in D \}$.
\end{proposition}

\begin{proof}
Let $X$ be the smallest superset of $W$ 
that is deductively closed 
and has the property that for any  $\alpha : \beta / \gamma \in D$,
if $\alpha\in X$, then $\gamma\in X$.
So, $X$ exists and is unique.
It remains to prove that $X$ is an extension of $(D,W)$.
Let $X = \cn(W\cup Y)$ where $Y$ is the set of all formulae introduced by means of the defaults in $D$.
So $Y \subseteq \{\beta\wedge\gamma \mid \alpha : \beta / \gamma \in D \}$.
Hence,  $\{\beta \mid \alpha : \beta / \gamma \in D \}$
is consistent with $\cn(W\cup Y)$.
So for any $\alpha : \beta / \gamma \in$, $\neg\beta\not\in X$.
Also if $\alpha\in X$, then $\gamma\in X$.
Therefore, we get the following applicability property: for any $\alpha : \beta / \gamma \in D$,
if $\alpha\in X$, and $\neg\beta\not\in X$, then $\gamma\in X$.
Since $X$ is the smallest set such that $W\subseteq X$,
and $\cn(X) = X$, and the applicability property holds,
then $X$ is an extension of $(D,W)$.
\end{proof}

The converse of the above proposition does not necessarily hold.
For instance, if $W$ is inconsistent, 
then $(D,W)$ is singular but $W$ is not consistent 
with $\{\beta\wedge\gamma \mid \alpha : \beta / \gamma \in D \}$,

More importantly for our purposes is that any default theory with an extension $E$ can be turned into a singular default theory by removing defaults without changing the extension $E$.

\begin{proposition}
For a default theory $(D,W)$, 
if $E$ is an extension of $(D,W)$, and $(D,W)$ is not singular, 
then there is a default theory $(D',W)$ s.t. $D' \subseteq D$ 
and $E$ is an extension of $(D',W)$, and $(D',W)$ is singular.
\end{proposition}

\begin{proof}
Either $\bot \in E$ or $\bot \not\in E$.
First assume $\bot \in E$.
But this holds if and only if $W\vdash\bot$.
Therefore let $D' = \emptyset$.
Therefore $(D',W)$ is singular.
Now assume $\bot \not\in E$.
Let $X$ = $\{ \alpha:\beta/\gamma \mid \alpha\in E \mbox{ and } \neg\beta\not\in E \}$.
Let $F(X) = \{ \gamma \mid \alpha:\beta/\gamma \in X\}$.
So $E = \cn(F(X)\cup W)$.
Let $D'$ be $X$. 
So $E$ is the unique extension of $(D',W)$.
\end{proof}

\begin{example}
Let $D = \{ {\tt a}:{\tt b}/{\tt b},  {\tt a}:\neg{\tt b}/\neg{\tt b} \}$ and $W = \{\tt a\}$.
So there are two extensions from $(D,W)$ which are $E_1 = \{  \tt a, b \}$
and $E_2 = \{  \tt a, \neg b \}$.
The subtheory $(D_1,W)$ where $D_1 = \{ {\tt a}:{\tt b}/{\tt b} \}$ is singular with the extension being $E_1$,
and the subtheory $(D_2,W)$ where $D_2 = \{ {\tt a}:\neg{\tt b}/\neg{\tt b} \}$ is singular with the extension being $E_2$.
\end{example}

So if $(D,W)$ is singular, and $E$ is the extension of $(D,W)$, 
then for every default in $\alpha:\beta/\gamma \in D$,
the default contributes to the extension (i.e. $\gamma \in E$),
or the default is not applied (i.e. $\alpha\not\in E$),
or the default can never be applied with $W$ (i.e. $W\cup\{\beta\}\vdash\bot$),
as illustrated in the following example.

\begin{example} 
\label{ex:singular:nonminimal}
Let $D = \{ {\tt a}:{\tt p}/{\tt q}, {\tt b}:\neg{\tt a}/{\tt r}, {\tt c}:{\tt s}/{\tt s} \}$ and $W = \{ \tt a, b \}$.
Therefore, $(D,W)$ is singular with the extension being $\cn(\{\tt a,b,q \})$
and where the first default (i.e. ${\tt a}:{\tt p}/{\tt q}$) is applied,
the second default (i.e. ${\tt b}:\neg{\tt a}/{\tt r}$) is not applied,
and the third default can never be applied with $W$ (i.e. ${\tt c}:{\tt s}/{\tt s}$).
\end{example}

For every default theory $(D,W)$ that is singular, 
it straightforward to identify a subset $D' \subseteq D$
such that $D'$ is a minimal set of defaults for each that extension of $D,W)$ is the same as the extension of $(D',W)$.
In this case, every default in $\alpha:\beta/\gamma \in D$ contributes to the extension (i.e. $\gamma \in E$),
as illustrated in the following example.

\begin{example} 
Continuing Example \ref{ex:singular:nonminimal},
let $D' = \{ {\tt a}:{\tt p}/{\tt q} \}$.
So $D' \subseteq D$ and every default in $D'$ contributes to the extension (i.e. ${\tt p}$ is in the extension of $(D',W)$). 
The extension of $(D',W)$ is $\cn(\{\tt a,b,q \})$,
and there is no subset $D''$ of $D$ such that the extension of $(D'',W)$ is $\cn(\{\tt a,b,q \})$.
\end{example}

Justifications from different default rules do not need to be consistent together for an extension to exist as illustrated by the following example. Note, in the situations where this kind of reasoning arises, we can consider counterarguments that can attack the reasoning (as we shall investigate later in the paper).

\begin{example} 
Let $D = \{ {\tt a}:{\tt p}/{\tt q}, {\tt b}:\neg{\tt p}/{\tt r} \}$ and $W = \{ \tt a, b \}$.
Therefore, $(D,W)$ is singular with the extension being $\cn(\{\tt a,b,q,r \})$. 
\end{example}

Some default theories have no extension as illustrated by the following example, and as such are not singular.

\begin{example}
(Example 2 in \cite{Froidevaux1994}) 
Let $D = \{ \top:{\tt a}/\neg{\tt a} \}$ and $W = \emptyset$.
There is no extension containing $\neg{\tt a}$ that is consistent with ${\tt a}$. 
\end{example}

We will use the definition of singular default theories in the definition of default argument that we introduce in the next section.

\section{Default arguments}
\label{section:defaultarguments}

We use the singular criteria (Definition \ref{def:singular}) in the following definition of a default argument to ensure that the implicit premises (respectively implicit claim) give a single perspective on the explicit premises (respectively explicit claim). 
The definition is very general, and we will consider constraints on this definition in order to give us appropriate notions of logical argument.

\begin{definition}
A {\bf default argument} is a tuple $\defaultargument$ 
where $\epset,\ecset \subseteq {\cal L}$, and $\ipset,\icset\subseteq{\cal D}$ s.t. 
$(\ipset,\epset)$ is singular
and 
$(\icset,\ecset)$ is singular.
\end{definition}

For a default argument $A = \defaultargument$,
we refer to $\epset$ as the {\bf explicit premises}, 
$\ipset$ as the {\bf implicit premises}, 
$\ecset$ as the {\bf explicit claims}, 
and $\icset$ as the {\bf implicit claims}. 
To extract these components of a default argument $A$,
we use the following functions:
$\ep(A) = \epset$;
$\ip(A) = \ipset$;
$\ec(A) = \ecset$;
and
$\ic(A) = \icset$.

\begin{example}
\label{ex:arguments}
The following are examples of default arguments:
\[
\begin{array}{l}
\tt A = \langle\tt  \{a\vee b\},\{(a\vee b\vee c:d/d)\}, \{d\}, \{(d:\neg e/\neg e)\}\rangle.\\
\tt B = \langle\tt \emptyset,\emptyset, \{b \vee \neg b\}, \emptyset\rangle.\\
\tt C = \langle\tt \emptyset,\{(:e/e)\}, \emptyset, \{(e:f/f), (f:g\wedge h/h)\}\rangle.\\
\end{array}
\]
\end{example}

As we will explore in the rest of this paper, a default argument provides a richer representation of an argument than available with other approaches to structured argumentation.
This includes the following features which go beyond other formalisms for logic-based argumentation:
{\bf Delineatation of implicit information connecting premises and claims:} 
The set $\ipset$ is a set of defaults that represents the implicit information in the premises;  
{\bf Logical mechanism for disabling connection between premises and claims:}
The justification of each default rule can be negated by the claim of another argument, thereby attacking the connection between the premise and claim;
{\bf Delineatation of implicit information connecting one argument with another:} 
The set $\icset$ is a set of defaults that represents the implicit information in the claim; 
{\bf Logical mechanism for disabling connection between one argument and another:}
The justification of each default rule can be negated by the claim of another argument, thereby attacking the connection between that argument and other arguments.

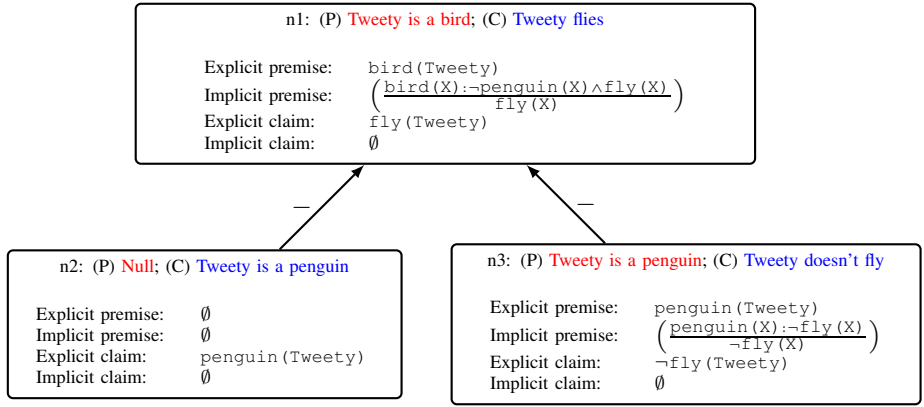
\begin{figure}[t]
\begin{center}
\begin{tikzpicture}[->,>=latex,thick, scale=0.8,
txtarg/.style={draw,text centered, text width=35mm,
shape=rectangle, rounded corners=2pt,
font=\scriptsize}]
\node (a1)[txtarg] [text width=80mm] at (4,4) {n1: (P) \myred{Tweety is a bird}; (C) \myblue{Tweety flies}
\[
\begin{array}{ll}
\mbox{Explicit premise: } & \mbox{\tt bird(Tweety) }\\
\mbox{Implicit premise: } & \left(\frac{\mbox{\tt bird(X)}:
\neg \mbox{\tt penguin(X)} \wedge \mbox{\tt fly(X)}}
{ \mbox{\tt fly(X)}} \right)\\
\mbox{Explicit claim: } & \mbox{\tt fly(Tweety)} \\
\mbox{Implicit claim: } &\emptyset\\
\end{array}
\]
};
\node[txtarg] (a2) [text width=50mm] at (0,0) {n2: (P) \myred{Null}; (C) \myblue{Tweety is a penguin}
\[
\begin{array}{ll}
\mbox{Explicit premise: }&\emptyset\\
\mbox{Implicit premise: }&\emptyset\\
\mbox{Explicit claim: } & \mbox{\tt penguin(Tweety) }\\
\mbox{Implicit claim: } &\emptyset\\
\end{array}
\]
};
\node[txtarg] (a3) [text width=60mm] at (8,0) {n3: (P) \myred{Tweety is a penguin}; (C) \myblue{Tweety doesn't fly}
\[
\begin{array}{ll}
\mbox{Explicit premise: }& \mbox{\tt penguin(Tweety) }\\
\mbox{Implicit premise: }& \left(\frac{\mbox{\tt penguin(X)}:
\neg \mbox{\tt fly(X)}}
{ \neg \mbox{\tt fly(X)}} \right)\\
\mbox{Explicit claim: } & \neg \mbox{\tt fly(Tweety) }\\
\mbox{Implicit claim: } & \emptyset\\
\end{array}
\]
};
\path	(a2) edge node[left] {$-$} (a1);
\path	(a3) edge node[right] {$-$} (a1);
\end{tikzpicture}
\end{center}
\caption{\label{fig:birdsfly}An example of an instantiated argument map concerning birds flying and penguins not flying. 
Here, there is no implicit claim in any of these arguments.
We explain these default arguments as follows:
Argument at n1 has an explicit premise of $\tt bird(Tweety)$ 
which satisfies the pre-condition for the default rule in the implicit premise,
and this gives the explicit claim $\tt fly(Tweety)$; 
Argument at n2 has the explicit claim of $\tt penguin(Tweety)$; 
and Argument at n3  has an explicit premise of $\tt penguin(Tweety)$ 
which satisfies the pre-condition for the default rule in the implicit premise, 
and this gives the explicit claim $\neg \tt fly(Tweety)$.
The explicit claim of arguments at n2 and n3 negate 
the justification condition for the default rule in the argument in node n1. 
Also, the claim of the argument at n3 negates the claim of the argument at n2.
}
\end{figure}

\begin{figure}[]
\begin{subfigure}{\textwidth}
\centering
\begin{tikzpicture}[->,>=latex,thick, scale=0.6,
txtarg/.style={draw,text centered, text width=35mm,
shape=rectangle, rounded corners=2pt,
fill=gray!20,font=\footnotesize},
]
\node[txtarg] (a1) [text centered,text width=8cm,shape=rectangle,draw] {n1: Patient has hypertension so prescribe diuretics.};
\node[txtarg] (a2) [below=of a1,text centered,text width=8cm,shape=rectangle,draw] {n2: Patient has hypertension so prescribe betablockers.};
\node[txtarg] (a3) [below=of a2,text centered,text width=8cm,shape=rectangle,draw] {n3: Patient has emphysema which is a contraindication for betablockers.};
\path	(a1)[bend left] edge node[right] {$-$} (a2);
\path	(a2)[bend left] edge node[left] {$-$} (a1);
\path	(a3) edge node[left] {$+$} (a2);
\end{tikzpicture}
\caption{\label{fig:medical:graph}An argument graph that captures the arguments and counterarguments in a decision making scenario.}
\end{subfigure}
\vspace{10mm}
\begin{subfigure}{\textwidth}
\centering
\begin{tikzpicture}[->,>=latex,thick, scale=0.6,
txtarg/.style={draw,text centered, text width=60mm,
shape=rectangle, rounded corners=2pt,
font=\scriptsize},
]
\node[] (x) at (0,15) {}; 
\node[txtarg] (a1) [text width=120mm] at (0,11) {n1: (P) \myred{Patient has hypertension}; (C) \myblue{Prescribe diuretics}
	\[
	\begin{array}{ll}
\mbox{Explicit premise: } & 	\mbox{\tt bloodpressure(high)}\\
\mbox{Implicit premise: } &    \left(\frac{\mbox{\tt bloodpressure(high): prescribe(diuretic}}{\mbox{\tt prescribe(diuretic)}} \right)\\
\mbox{Explicit claim: }	& \mbox{\tt prescribe(diuretic)}		\\
\mbox{Implicit claim: }	& \left(\frac{\mbox{\tt prescribe(diuretic)}: \neg \mbox{\tt prescribe(betablocker)}}{ \neg \mbox{\tt prescribe(betablocker)}} \right)\\
	\end{array}
	\]};
\node[txtarg] (a2) [text width=120mm] at (0,5) {n2: (P) \myred{Patient has hypertension}; (C) \myblue{Prescribe betablockers}
	\[
	\begin{array}{ll}
\mbox{Explicit premise: } &	\mbox{\tt bloodpressure(high)} \\
\mbox{Implicit premise: } &     \left(\frac{\mbox{\tt bloodpressure(high): prescribe(betablocker}}{\mbox{\tt prescribe(betablocker)}} \right)\\	
	\mbox{Explicit claim: }	& \mbox{\tt prescribe(betablocker)}		\\
 \mbox{Implicit claim: }	& \left(\frac{\mbox{\tt prescribe(betablocker)}: \neg \mbox{\tt prescribe(diuretic)}}{ \neg \mbox{\tt prescribe(diuretic)}} \right)\\
	\end{array}
	\]};
\node[txtarg] (a3) [text width=120mm] at (0,0) {n3: (P) \myred{Patient has emphysema}; (C) \myblue{Contraindication for betablockers}
\[
\begin{array}{ll}
\mbox{Explicit premise: } &\mbox{\tt symptom(emphysema)}, \\
\mbox{Implicit premise: } & \left(\frac{\mbox{\tt symptom(emphysema)}:\mbox{\tt contraindication(betablocker)}}{ \mbox{\tt contraindication(betablocker)}} \right)\\
\mbox{Explicit claim: }	& \mbox{\tt contraindication(betablocker)}\\
\mbox{Explicit premise: } & \left(\frac{\mbox{\tt contraindication(betablocker)}: \neg \mbox{\tt prescribe(betablocker)}}{ \neg \mbox{\tt prescribe(betablocker)}} \right)\\
\end{array}
\]};
\path	(a1)[bend left] edge node[right] {$-$} (a2);
\path	(a2)[bend left] edge node[left] {$-$} (a1);
\path	(a3) edge node[left] {$+$} (a2);
\end{tikzpicture}
\caption{\label{fig:medical:logic}An instantiated argument map generated from the argument graph in Figure \ref{fig:medical:graph}.}
\end{subfigure}
\caption{\label{fig:medical}A decision making scenario where there are two alternatives for treating a patient, diuretics or betablockers. Since only one treatment should be given for the disorder, each argument attacks the other. There is also a reason to not give betablockers, as the patient has emphysema which is a contraindication for this treatment.}
\end{figure}
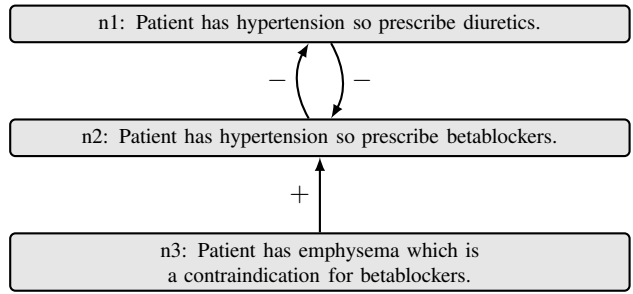
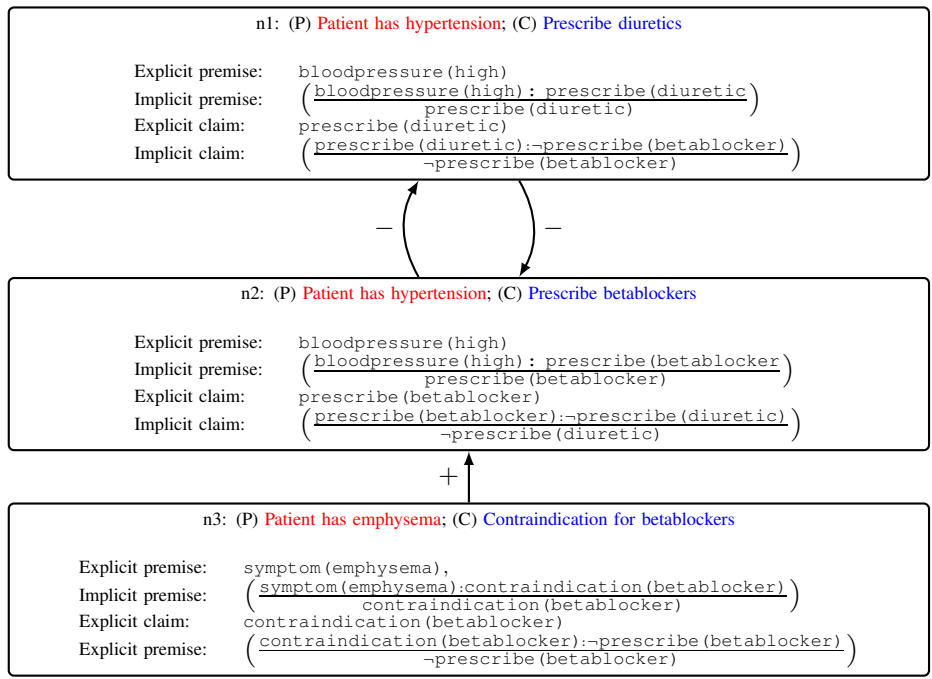

We give further examples of default arguments in each node in the instantiated argument maps 
in Figure \ref{fig:birdsfly}, Figure \ref{fig:intrograph3}, and Figure \ref{fig:medical}:
Figure \ref{fig:birdsfly} provides a straightforward and common kind of example to illustrate how it can be handled in our approach; 
and
Figure \ref{fig:intrograph3} captures the argument map from Figure \ref{fig:intrograph2}, and it shows the value of implicit claims --- in particular, it shows how implict claims capture how one enthymeme attacks or supports another enthymeme.
We will explain  the instantiation process used in these examples in more detail in later sections including the nature of attacks and support for default arguments.

\begin{figure}
\begin{center}
\begin{tikzpicture}
[->,>=latex,thick, scale=0.6,
txtarg/.style={draw,text centered, 
shape=rectangle, rounded corners=2pt,
font=\footnotesize}
]
\node[txtarg] (a0) [text width=90mm] at (0,24) {n0: (P) \myred{Null}; (C) \myblue{We should ban cars from cities}
\[
\begin{array}{ll}
\mbox{Explicit premise: } & 	\emptyset\\
\mbox{Implicit premise: } &    \{ \left(\frac{ \mbox{\tt $\top$}:\mbox{\tt s0}}{\mbox{\tt s0}} \right)
                                \left(\frac{ \mbox{\tt s0}:\mbox{\tt s1}}{\mbox{\tt s1}} \right) \}\\
\mbox{Explicit claim: }	& \mbox{\tt s1}\\
\mbox{Implicit claim: }	& \emptyset\\
	\end{array}
\]
};
\node[txtarg] (a1) [text width=90mm] at (0,12) {n1: (P) \myred{Cars are polluting}; (C) \myblue{Cars are bad for the health}
\[
\begin{array}{ll}
\mbox{Explicit premise: } & 	\mbox{\tt s3}\\
\mbox{Implicit premise: } &  \{ \left(\frac{ \mbox{\tt s3}:\mbox{\tt s0}}{\mbox{\tt s0}} \right) \}\\
\mbox{Explicit claim: }	& \mbox{\tt s0}\\
\mbox{Implicit claim: }	& \emptyset\\
\end{array}
\]
};
\node[txtarg] (a2) [text width=90mm] at (9,18) {n2: (P) \myred{Null}; (C) \myblue{Cars are vital for people to move around cities}
\[
\begin{array}{ll}
\mbox{Explicit premise: } & 	\emptyset\\
\mbox{Implicit premise: } &  \{ \left(\frac{ \top : \mbox{\tt s2} }{\mbox{\tt s2}} \right)\}\\
\mbox{Explicit claim: }	& \mbox{\tt s2}\\
\mbox{Implicit claim: }	& \{ \left(\frac{ \mbox{\tt s2} : \neg \mbox{\tt s1}}{\neg \mbox{\tt s1}} \right)\}\\
\end{array}
\]
};
\node[txtarg] (a3) [text width=90mm] at (0,0) {n3: (P) \myred{Null}; (C) \myblue{Internal combustion engines pollute}
\[
\begin{array}{ll}
\mbox{Explicit premise: } & 	\emptyset\\
\mbox{Implicit premise: } & \{ \left(\frac{ \top:\mbox{\tt s4}}{\mbox{\tt s4}} \right)\} \\
\mbox{Explicit claim: }	& \mbox{\tt  s4}\\
\mbox{Implicit claim: }	& \{ \left(\frac{ \mbox{\tt s4}:\mbox{\tt s3}}{\mbox{\tt s3}} \right) \}\\
\end{array}
\]
};
\node[txtarg] (a4) [text width=90mm] at (9,6) {n4: (P) \myred{Legislation}; (C) \myblue{Soon all new cars will be electric}
\[
\begin{array}{ll}
\mbox{Explicit premise: } & 	\mbox{\tt  s5}\\
\mbox{Implicit premise: } &  \{ \left(\frac{ \mbox{\tt s5}:\mbox{\tt s6}}{\mbox{\tt s6}} \right) \}\\
\mbox{Explicit claim: }	& \mbox{\tt  s6}\\
\mbox{Implicit claim: }	& \{ \left(\frac{ \mbox{\tt s6}:\neg \mbox{\tt s3}}{\neg \mbox{\tt s3}} \right) \}\\
\end{array}
\]
};
\path	(a1) edge node[left] {$+$} (a0);
\path	(a2) edge node[right] {$-$} (a0);
\path	(a3) edge node[left] {$+$} (a1);
\path	(a4) edge node[right] {$-$} (a1);
\end{tikzpicture}
\end{center}
\caption{\label{fig:intrograph3}An instantiated argument map for the argument map in Figure \ref{fig:intrograph2}.
The atoms are ${\tt s0}$ = {\em cars are bad for health},
${\tt s1}$ = {\em Cars should be banned from cities},
${\tt s2}$ = {\em Cars are vital for people to move around cities},
${\tt s3}$ = {\em Cars are polluting},
${\tt s4}$ = {\em Internal combustion engines pollute},
${\tt s5}$ = {\em Legislation}, 
and ${\tt s6}$ = {\em Soon all new cars will be electric}.
We explain the structuring of the arguments at each node as follows:
(n0) argument has implicit premises that imply the explicit claim ${\tt s1}$;
(n1) argument has the explicit premise ${\tt s3}$, which is used to derive the explicit claim ${\tt s0}$, and this is a support for argument at n0;
(n2) argument has the implicit premise that is used to derive the explicit claim ${\tt s2}$, and this is then used with the implicit claim to give $\neg {\tt s1}$, and this constitutes an attack on argument at n0;
(n3) argument has implicit premises that is used to derive $\tt s3$, and this is then used as a support for the explicit premises of the argument at n1;
and 
(n4) argument has the explicit claim $\neg \tt s3$, and this is then used as an attack on the explicit premise of the argument at n1.
}
\end{figure}
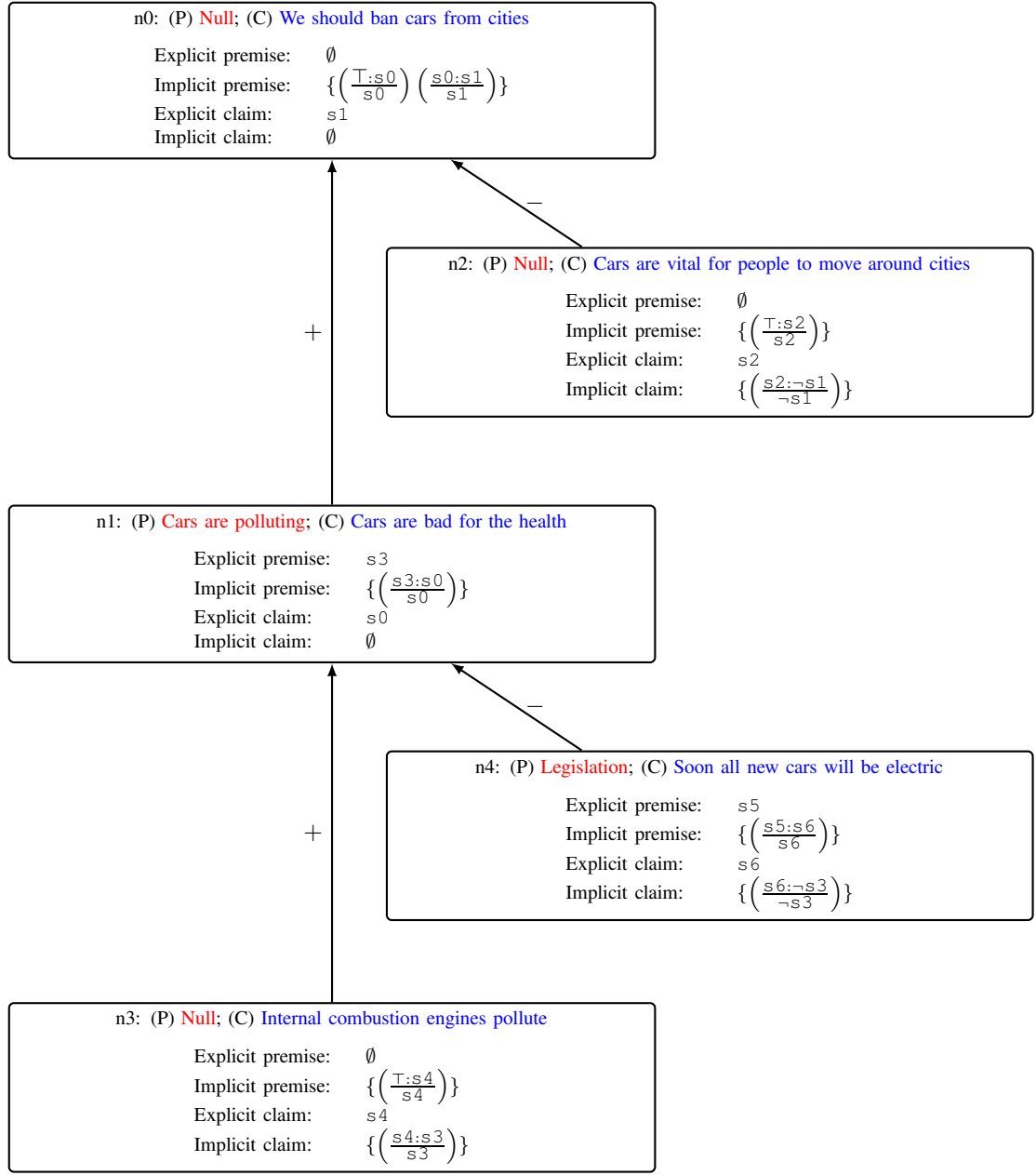

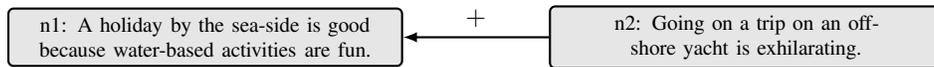
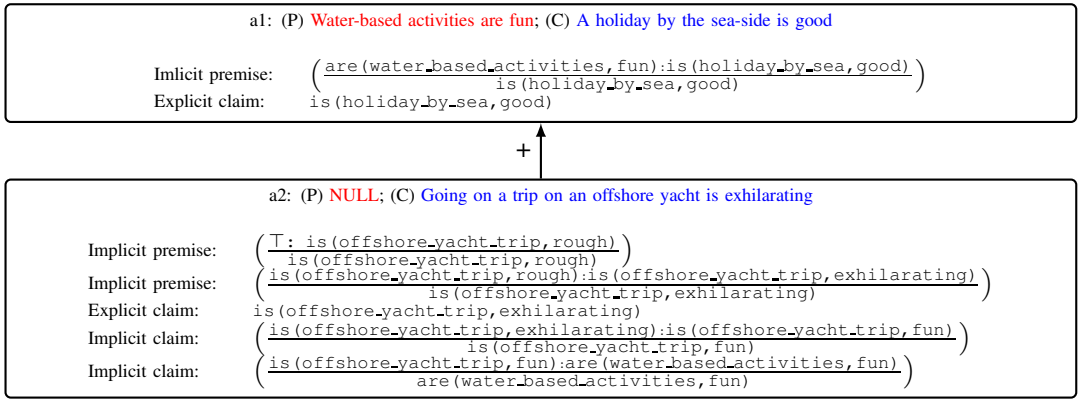
\begin{figure}
\begin{subfigure}{\textwidth}
\centering
\begin{tikzpicture}
[->,>=latex,thick, scale=0.6,
txtarg/.style={draw,text centered, 
shape=rectangle, rounded corners=2pt,
fill=gray!20,font=\footnotesize}]
\node[txtarg] (a1) [text width=50mm] at (0,0) {n1: A holiday by the sea-side is good because water-based activities are fun.};
\node[txtarg] (a2) [text width=50mm] at (12,0)  {n2: Going on a trip on an offshore yacht is exhilarating.};
\path	(a2) edge node[above] {$+$} (a1);
\end{tikzpicture}
\caption{\label{fig:seaside:map}An argument graph where the argument at n2 is a supporting argument for the argument at n1.}
\end{subfigure}
\begin{subfigure}{\textwidth}
\centering
\begin{tikzpicture}[->,>=latex,thick, scale=0.6,
txtarg/.style={draw,text centered, text width=35mm,
shape=rectangle, rounded corners=2pt,
font=\scriptsize}]
\node[] (x) at (0,10) {};
\node[txtarg] (a1) [text width=140mm] at (0,5) {a1: (P) \myred{Water-based activities are fun}; (C) \myblue{A holiday by the sea-side is good}
	\[
	\begin{array}{ll}
\mbox{Imlicit premise: } &   \left(\frac{\mbox{\tt are(water\_based\_activities,fun)}:
        \mbox{\tt is(holiday\_by\_sea,good)}}
        {\mbox{\tt is(holiday\_by\_sea,good)}} \right)\\
\mbox{Explicit claim: } &	\mbox{\tt  is(holiday\_by\_sea,good)}					
	\end{array}
	\]};
\node[txtarg] (a2) [text width=140mm] at (0,0) {a2: (P) \myred{NULL}; (C) \myblue{Going on a trip on an offshore yacht is exhilarating}
	\[
	\begin{array}{ll}
\mbox{Implicit premise: } &       \left(\frac{\mbox{\tt $\top$: is(offshore\_yacht\_trip,rough)}}{\mbox{\tt              is(offshore\_yacht\_trip,rough)}} \right)\\
\mbox{Implicit premise: } &       \left(\frac{\mbox{\tt is(offshore\_yacht\_trip,rough)}:
            \mbox{\tt is(offshore\_yacht\_trip,exhilarating)}}
        {\mbox{\tt is(offshore\_yacht\_trip,exhilarating)}} \right)\\
\mbox{Explicit claim: } &   \mbox{\tt is(offshore\_yacht\_trip,exhilarating)}\\
\mbox{Implicit claim: } &    \left(\frac{\mbox{\tt is(offshore\_yacht\_trip,exhilarating)}:
            \mbox{\tt is(offshore\_yacht\_trip,fun)}}
        {\mbox{\tt is(offshore\_yacht\_trip,fun)}} \right)\\
\mbox{Implicit claim: } &    \left(\frac{\mbox{\tt is(offshore\_yacht\_trip,fun)}:
        \mbox{\tt are(water\_based\_activities,fun)}}
        {\mbox{\tt are(water\_based\_activities,fun)}} \right)\\
	\end{array}
	\]};
\path	(a2)[] edge node[left] {+} (a1);
\end{tikzpicture}
\caption{\label{fig:seaside:logic}An instantiated argument map generated from the argument map in Figure \ref{fig:seaside:map}.}
\end{subfigure}
\caption{\label{fig:seaside}An example of an instantiated argument map in Figure \ref{fig:seaside:logic} generated from the argument map, concerning a type of holiday, in Figure \ref{fig:seaside:map}. In this example, we see the use of the implicit claim to generate an inference in the consequence of the supporting argument that is a fact in the premises of the supported argument.}
\end{figure}

Let $\cal R$ be the set of all possible default rules.
The {\bf set of default arguments} from a set of default rules $\Pi\subseteq{\cal R}$ is
${\sf Arguments}(\Pi) 
= \{ \defaultargument \mid \epset,\ecset \subseteq{\cal L} 
\mbox{ and } \ipset,\icset \subseteq \Pi \}$.
Using default logic as a base logic in this way does not affect the argument construction being monotonic; Adding a formula to the knowledge-base would not cause a default argument to be withdrawn. Rather it may allow further default arguments to be constructed. 
So if $\Pi\subseteq\Pi'$,
then ${\sf Arguments}(\Pi) \subseteq {\sf Arguments}(\Pi')$.

Given a default argument $A$,
the support of the argument is the default extension obtained from the implicit and explicit premises,
and the consequence of the argument is the default extension obtained from the implicit and explicit claim:
\begin{itemize}[noitemsep]
\item The {\bf support} of $A$ is $\support(A) = \ex(\ip(A),\ep(A))$.
\item The {\bf consequence} of $A$ is $\consequence(A) = \ex(\ic(A),\ec(A))$.
\end{itemize}

So the support of a default argument captures the consequences of the implicit and explicit premises 
and the claim of a default argument captures the consequences of the implicit and explicit claims.

\begin{example}
Continuing Example \ref{ex:arguments}, 
for the default argument $\tt A$ to $\tt C$, we have the following support and consequence.
\[
\begin{array}{lll}
\tt A = \langle\tt \{ a\vee b \},\{(a\vee b\vee c:d/d)\}, \{d\}, \{(d:\neg e/\neg e)\}\rangle
& \support({\tt A}) = \cn(\{\tt a\vee b, d \})
& \consequence({\tt A}) = \cn(\{\tt d, \neg e \})\\
\tt B = \langle\tt \emptyset,\emptyset, \{b \vee \neg b\}, \emptyset\rangle
& \support({\tt B}) = \cn(\emptyset)
& \consequence({\tt B}) = \cn(\emptyset)\\
\tt C = \langle\tt \emptyset,\{(\top:e/e),(f:g\wedge h/h)\}, \{h\}, \{(h:f/f)\}\rangle
& \support({\tt C}) = \cn(\{\tt e \})
& \consequence({\tt C}) = \cn(\{\tt f, h \})\\
\end{array}
\]
\end{example}

The definition for a default argument is very general, 
and so we consider some types of argument including the following.

\begin{definition}
For a default argument $A$, 
\begin{itemize}[noitemsep]
\item $A$ is {\bf valid} iff $\ec(A) \subseteq \support(A)$. 
\item $A$ is {\bf implicitly minimal} 
iff $A$ is valid and there is no $\Phi\subset\ip(A)$ s.t. $\ec(A) \subseteq \ex(\Phi,\ep(A))$.
\item $A$ is {\bf explicitly minimal} 
iff $A$ is valid and there is no $\Psi\subset\ep(A)$ s.t. $\ec(A) \subseteq \ex(\ip(A),\Psi)$.
\item $A$ is {\bf support consistent} 
iff $\bot \not\in\support(A)$.
\item $A$ is {\bf consequence consistent} 
iff $\bot \not\in\consequence(A)$.
\item $A$ is {\bf fully consistent} 
iff $\support(A)\cup\consequence(A)\not\vdash\bot$.
\end{itemize}
\end{definition}

We explain the above definitions as follows:
A default argument is valid iff the explicit claim is in the support of the argument (i.e. the extension of the premises);
A default argument is implicitly minimal iff there is no subset of the implicit premises that is valid;
A default argument is explicitly minimal iff there is no subset of the explicit premises that is valid;
A default argument is support consistent iff the support is consistent;
A default argument is claim consistent iff the claim is consistent;
And a default argument is fully consistent iff the support and claim are consistent together.
Clearly, if an argument is fully coherent, then it is support consistent and claim consistent.

\begin{example}
Continuing Example \ref{ex:arguments},
default arguments $\tt A$. $\tt B$, and $\tt C$, are valid, fully consistent,
and minimal,
whereas $\tt D$ is valid and fully consistent but not minimal. 
\[
\begin{array}{l}
\tt A = \langle\tt  \{a\vee b\},\{(a\vee b\vee c:d/d)\}, \{d\}, \{(d:\neg e/\neg e)\}\rangle.\\
\tt B = \langle\tt \emptyset,\emptyset, \{ b \vee \neg b\}, \emptyset\rangle.\\
\tt C = \langle\tt \emptyset,\{(:e/e)\}, \{e\}, \{(e:f/f), (f:g\wedge h/h)\}\rangle.\\
\tt D = \langle\tt \emptyset,\{(:e/e)\}, \emptyset, \{(e:f/f), (f:g\wedge h/h)\}\rangle.\\
\end{array}
\]
\end{example}

\begin{example}
The default arguments $\tt A = \langle \{a\},\{(a:b/b)\}, \{b\},\{ (b:\neg a/\neg a)  \}\rangle$
and $\tt B = \langle \{a\},\emptyset,\{\neg a\},\emptyset\rangle$ 
are support and consequence consistent (as shown below) but they are not fully consistent since 
$\support(A) \cup  \consequence(A) \vdash \bot$
and $\support(B) \cup  \consequence(B) \vdash \bot$. 
\[
\begin{array}{cc}
\support(A) = \cn(\{\tt a,b \})   &  \consequence(A) = \cn(\{\tt \neg a, b \})\\
\support(B) = \cn(\{\tt a\})   &  \consequence(B) = \cn(\{\tt \neg a\})\\
\end{array}
\]
\end{example}

A default argument $A$ is {\bf explicit} iff $\ip(A) = \ic(A) = \emptyset$.
If $A$ is explicit, valid, and fully consistent, 
then for each $\beta\in\ec(A)$, $\ep(A)\vdash\beta$.
A default argument $A$ is a {\bf classical argument}
iff $A$ is explicit, valid, explicitly minimal, and support consistent.
So a classical argument is a minimal argument,
and a classical argument is fully consistent. 
A default argument $A$ is a {\bf non-classical argument} 
iff $A$ is not a classical argument.

\begin{example}
The default argument $\langle\tt \{a \vee b\},\{(a\vee b\vee c:d/d)\},\{a\vee b \vee e\},\emptyset\rangle$
is valid, explicitly minimal, and fully consistent, but it is not explicit nor implicitly minimal, 
and therefore it is not classical, 
whereas the default argument $\langle\tt \{a \vee b\},\emptyset,\{a\vee b \vee e\},\emptyset\rangle$
is explicit, explicitly minimal, valid, and fully consistent,
and therefore classical. 
The following default arguments are non-classical:
$\langle\tt \{a\},\{(a:b/c)\},\{c\},\emptyset\rangle$, 
$\langle\tt \{a\},\{(a:b/c)\},\{d\},\emptyset\rangle$, 
$\langle\tt \{a\},\emptyset,\{d\},\emptyset\rangle$,
and 
$\langle\tt \{a\wedge\neg a\},\emptyset,\{d\},\emptyset\rangle$
\end{example}

We refer to argument $\langle\emptyset,\emptyset,\emptyset,\emptyset\rangle$ as a {\bf vacuous argument}.
It is valid, implicitly minimal, explicitly minimal, fully consistent, and classical.

\begin{example}
Continuing Example \ref{ex:arguments},
the default argument 
$\tt B = \langle\tt \emptyset,\emptyset, \{b \vee \neg b\}, \emptyset\rangle$ is vacuous.
\end{example}

A default argument $A$ has {\bf completely implicit premises} 
iff $\ep(A) = \emptyset$ and $\ip(A) \neq \emptyset$;
and 
$A$ has a {\bf completely implicit claims} 
iff $\ec(A) = \emptyset$ and $\ic(A) \neq \emptyset$.

\begin{example}
Continuing Example \ref{ex:arguments},
the argument 
$\tt C = \langle\tt \emptyset,\{(:e/e)\}, \emptyset, \{(e:f/f), (f:g\wedge h/h)\}\rangle$
has completely implicit premises and claims,
whereas $\tt D = \langle\tt \emptyset,\{(:e/e)\}, \{e\}, \emptyset\rangle$
has completely implicit premises and explicit claims.
\end{example}

\begin{proposition}
For all $\Pi\subseteq{\cal R}$,
where ${\cal R}$ is the set of default rules, 
if $A$ is a classical argument, 
then $A\in {\sf Arguments}(\Pi)$.
\end{proposition}

\begin{proof}
If $A$ is a classical argument, 
then it is of the form $\defaultargument$.
Let $\ipset = \icset = \emptyset$,
and let $\epset$ and $\ecset$ are such that 
for each $\beta\in\ecset$,
$\epset\vdash\beta$.
So $A\in {\sf Arguments}(\Pi)$.
\end{proof}

Default logic does not allow an inconsistent extension except when the premise is inconsistent.
So an inconsistent claim arises only if the explicit premise is inconsistent.

\begin{proposition}
If default argument $A$ is a valid explicit argument,
and $\ec(A)\vdash\bot$,
then $\bot\in\ep(A)$.
\end{proposition}

\begin{proof}
Assume $A$ is explicit.
So $\ip(A) = \emptyset$ and $\ic(A) = \emptyset$.
Also assume $A$ is valid. 
So $\ec(A) \subseteq \ex(\ip(A),\ep(A))$.
Recall that $\ec(A)\vdash\bot$ is assumed. 
Now the only way for $\bot \in \ex(\ip(A),\ep(A))$ to hold,
is for $\bot\in\ep(A)$ to hold.
\end{proof}

For a default argument $A$,
if neither $\ep(A)$ nor $\ec(A)$ is contradictory, 
then $A$ is consistent, as captured by the following result.

\begin{proposition}
For default argument $A$,
(1) if $\bot\in\support(A)$, 
then $\ep(A)\equiv\bot$;
(2) if $\bot\in\consequence(A)$, 
then $\ec(A)\equiv\bot$.
\end{proposition}

\begin{proof}
(1) Assume $\bot\in\support(A)$.
So $\ex(\ip(A),\ep(A))$ is inconsistent.
But this is only possible if $\ep(A)$ is inconsistent.
So $\ep(A)\equiv\bot$ holds.
(2) Assume $\bot\in\consequence(A)$.
So $\ex(\ic(A),\ec(A))$ is inconsistent.
But this is only possible if $\ec(A)$ is inconsistent.
So $\ec(A)\equiv\bot$ holds.
\end{proof}

\begin{example}
Consider the following arguments.
\begin{itemize}[noitemsep]
\item $A =  \tt \langle \{a\wedge\neg a\}, \emptyset, \{d\}, \{ d:e/e \} \rangle$ is such that $\bot\in\support(A)$ and $\bot\not\in\consequence(A)$.
\item $B =  \tt \langle \{a\}, \{ a:b/b\}, \{d\wedge \neg d\}, \emptyset \rangle$ 
is such that $\bot\not\in\support(B)$ and $\bot\in\consequence(B)$.
\end{itemize}
\end{example}

With the richer structure that comes with default arguments, there are various ways that two arguments can be formalized as equivalent. 

\begin{definition}
Types of equivalence between default arguments $A$ and $B$ include:
\begin{itemize}[noitemsep]
\item $A$ is {\bf explicitly equivalent} to $B$ 
iff $\cn(\ep(A)) = \cn(\ep(B))$ 
and $\cn(\ec(A)) = \cn(\ec(B))$.
\item $A$ is {\bf support equivalent} to $B$ 
iff $\support(A) = \support(B)$.
\item $A$ is {\bf consequence equivalent} to $B$ 
iff $\consequence(A) = \consequence(B)$.
\item $A$ is {\bf implicitly equivalent} to $B$ 
iff $A$ is support equivalent 
and consequence equivalent to $B$.
\item $A$ is {\bf intrinsically equivalent} to $B$ 
iff $A$ is support equivalent to $B$
and $\cn(\ec(A)) = \cn(\ec(B))$.
\end{itemize}
\end{definition}

We explain the above definitions as follows:
$A$ is explicitly equivalent to $B$ when the explicit premises are equivalent and the explicit claims are equivalent; 
$A$ is support equivalent to $B$ when the supports are the same;
$A$ is claim equivalent to $B$ when the claims are the same;
and $A$ is implicitly equivalent to $B$ when they are both premise and claim equivalent.
Since the implicitness of an argument's claim depends on its relations to another argument, $A$ is intrinsically equivalent to $B$ when they are both equivalent regardless of their relations.

\begin{example}
Default arguments $\tt A$, $\tt B$ and $\tt C$ are implicitly equivalent.
\[
\begin{array}{l}
\tt A = \langle\tt \{e\},\{(e:f/f)\}, \{f\}, \{(f:\neg b/\neg b)\}\rangle.\\
\tt B = \langle\tt \{e\},\{(e:f/f)\}, \{f\wedge \neg b\}, \emptyset\rangle.\\
\tt C = \langle\tt \emptyset,\{(:e/e),(e:f/f)\}, \{\neg (\neg f\vee b)\}, \emptyset\rangle.\\
\end{array}
\]
\end{example}

\begin{proposition}
For every classical (respectively valid, implicitly minimal, and explicitly minimal, non-classical) argument $A$,
there is a  valid, implicitly minimal, and explicitly minimal, non-classical (respectively classical) argument $B$,
s.t. $A$ is implicitly equivalent to $B$.
\end{proposition}

\begin{proof}
Let $A$ be a classical argument.
So $\ip(A) = \emptyset$ and $\ic(A) = \emptyset$.
Let $B$ be a valid, implicitly minimal, and explicitly minimal, non-classical argument.
So $\ip(B) \neq \emptyset$ or $\ic(B) \neq \emptyset$. 
Therefore, for each $A$, a $B$ can be chosen 
so that $\support(A) \equiv \support(B)$ 
and $\consequence(A) \equiv \consequence(B)$,
and hence so that $A$ is implicitly equivalent to $B$.
In the same way, for each $B$, an $A$ can be chosen 
so that they are implicitly equivalent.
\end{proof}

We can also rank arguments by their implicitness: $A$ is more implicit than $B$ when they are implicitly equivalent but the explicit premises and claim of $A$ are weaker than those of $B$.

\begin{itemize}
 \item $A$ is {\bf more implicit} than $B$ 
iff $A$ is implicitly equivalent to $B$ 
and $\cn(\ep(A)) \subseteq \cn(\ep(B))$
and $\cn(\ec(A)) \subseteq \cn(\ec(B))$.
\end{itemize}

To conclude this section, default arguments are a general kind of argument, with some important special cases, that meet our needs for formalizing textual arguments as arising in argument maps. The advantage of using default logic in arguments is that it allows default inferences to be drawn. 
This means that we use a well-developed and well-understand formalism for representing and reasoning with the complexities of default knowledge. Hence, we can have a richer and more natural representation of defaults. 
It also means that inferences can be drawn in the absence of reasons to not draw them. For instance, we can conclude the something flies from knowing that is a bird in the absence of knowing whether it is a penguin. In other words, we just need to know that it is consistent to believe that it is not a penguin and that it is consistent to believe that it can fly. Furthermore, we just need to do this consistency check within the premises of the argument. Note, this consistency check is different to the consistency check used for default negation in DeLP which involves checking consistency with all the strict knowledge (i.e. the subset of knowledge that is assumed to be correct) \cite{GS04}.

\section{Relationships between arguments}

We now consider how one default argument supports or attacks another default argument. 
For this, we will require some subsidiary definitions. 
The first is to identify the justifications that arise in the default rules in the premises of an argument using the following definition for the {\bf premise justification function}, where $A$ is an argument.
\[
\justification(A) = \{ \beta \mid \alpha:\beta/\gamma \in \ip(A) \}
\]

\begin{example}
For $\tt A = \langle\tt \{e\},\{(e:f\wedge a/f)\}, \{f\}, \{(f:\neg b \wedge (a \vee c)/\neg b)\}\rangle$, 
$\justification(\tt A) = \{\tt f\wedge a \}$.
\end{example}

We also require the following variants of the $\support$ and $\consequence$ functions. 
For any default argument $A$, $\consequence^*(A) = \consequence(A)\setminus\cn(\{\top\})$
and $\support^*(A) = \support(A)\setminus\cn(\{\top\})$.
We use $\consequence^*(A)$ and  $\support^*(A)$ instead of $\consequence(A)$ and  $\support(A)$ as we want to avoid trivial support involving tautologies. 

We define the following notions of support relation that hold between a pair of default arguments. 
Whilst there are further kinds of support that we could define, we start with these intuitive options.
Essentially, we restrict the definitions to comparing the explicit claim or consequence (i.e. extension of the implicit and explicit claims) of a supporting argument $A$ (i.e. $\ec(A)$ or $\consequence(A)$), 
and the explicit premises, or support (i.e. extension of the implicit and explicit premises), or justification, of the supported argument $B$
(i.e. $\ep(B)$, or $\support(B)$, or $\justification(B$)), and we restrict the comparison between the supporting and supported arguments to an intersection between the respective sets of formulae.

\begin{definition}
\label{def:support}
For default arguments $A$ and $B$,
the following are types of support.
\begin{itemize}[noitemsep]
\item $A$ {\bf inferentially supports} $B$ 
iff  $\support^*(B)\cap\consequence^*(A)\neq\emptyset$.
\item $A$ {\bf directly supports} $B$ 
iff  $\ep(B)\cap\consequence^*(A)\neq\emptyset$.
\item $A$ {\bf explicitly supports} $B$ 
iff  $\ep(B)\cap\cn(\ec(A))\neq\emptyset$.
\item $A$ {\bf justification supports} $B$ 
iff  $\justification(B)\cap\consequence^*(A)\neq\emptyset$.
\end{itemize}
\end{definition}

We explain the above definitions as follows and provide an example below:
Inferential support ensures that there is a consequence of $A$ that is a support in $B$;
Direct support ensures that there is a consequence of $A$ that is an explicit premise in $B$;
Explicit support ensures that the explicit claim in $A$ implies an explicit premise in $B$; 
And justification support ensures that there is a consequence of $A$ that is a justification in the implicit premises in $B$.

\begin{example}
Consider the following supporting default arguments (left) and supported default arguments (right):
$\tt A1$ inferentially, directly, but not explicitly, supports $\tt B1$.
$\tt A2$ inferentially, directly, and explicitly, supports $\tt B2$.
$\tt A3$ justification, but not inferentially, supports $\tt B3$.
$\tt A4$ justification, inferentially, directly, and explicitly, supports $\tt B4$.
\[
\begin{array}{ll}
\tt A1 = \langle\tt \emptyset,\emptyset,\{a\},\{(a:b/b)\}\rangle
&\tt  B1 = \langle\tt \{b\},\{ (b:c/c) \},\{c\},\emptyset\rangle\\
\tt A2 = \langle\tt \emptyset,\emptyset,\{a\},\emptyset\rangle
&\tt  B2 = \langle\tt \{a\},\{ (a:b/c) \},\{c\},\emptyset\rangle\\
\tt A3 = \langle\tt \emptyset,\emptyset,\{b\},\emptyset\rangle
&\tt  B3 = \langle\tt \{a\},\{ (b:c/c) \},\{c\},\emptyset\rangle\\
\tt A4 = \langle\tt \emptyset,\emptyset,\{a\wedge b\},\emptyset\rangle
&\tt  B4 = \langle\tt \{a\},\{ (a:b/c) \},\{c\},\emptyset\rangle\\
\end{array}
\]
\end{example}

For each form of support above, we provide a more restricted form of the support. Essentially, above the relationship between the formulae considered in the supported argument and supporting argument is intersection, whereas below we replace intersection with the subset or equal relation. In each definition, we recall the more general definition to ensure that there is a non-empty set of formulae considered for the supported and supporting arguments, otherwise we may have trivial support. For example, for inferentially supports, if $\support(B)$ is the emptyset, and we did not specify that $A$ inferentially supports $B$ holds, then we would have that $A$ strongly inferentially supports $B$, and so we would have trivial form of support. Ensuring that  $A$ inferentially supports $B$ holds obviates this possibility. 

\begin{definition}
\label{def:strongsupport}
For default arguments $A$ and $B$,
the following are types of support.
\begin{itemize}[noitemsep]
\item $A$ {\bf strongly inferentially supports} $B$ 
iff  $A$ inferentially supports $B$ 
and $\support(B)\subseteq\consequence(A)$.
\item $A$ {\bf strongly directly supports} $B$ 
iff  $A$ directly supports $B$
and $\ep(B)\subseteq\consequence(A)$.
\item $A$ {\bf strongly explicitly supports} $B$ 
iff  $A$ explicitly supports $B$
and $\ep(B)\subseteq\cn(\ec(A))$.
\item $A$ {\bf strongly justification supports} $B$ 
iff  $A$ justification supports $B$ 
and $\justification(B)\subseteq\consequence(A)$.
\end{itemize}
\end{definition}

\begin{example}
Consider the following supporting default arguments (left) and supported default arguments (right):
$\tt A1$ strongly inferentially, but not strongly directly, nor strongly explicitly, supports $\tt B1$.
$\tt A2$  strongly directly, but not strongly explicitly, nor strongly inferentially, supports $\tt B2$.
$\tt A3$ strongly directly, and strongly explicitly, but not strongly inferentially, supports $\tt B3$.
$\tt A4$ strongly inferentially, strongly directly, and strongly explicitly,  supports $\tt B4$.
and $\tt A5$ strongly justification, but not strongly inferentially, nor strongly directly, nor strongly explicitly, supports $\tt B4$.
\[
\begin{array}{ll}
\tt A1 = \langle\tt \emptyset,\emptyset,\{a\},\{(a:b/b)\}\rangle
&\tt  B1 = \langle\tt \emptyset,\{ (\top:b/b) \},\{b\},\emptyset\rangle\\
\tt A2 = \langle\tt \emptyset,\emptyset,\{a\},\{(a:b/b)\}\rangle
&\tt  B2 = \langle\tt \{b\},\{ (b:c/c) \},\{c\},\emptyset\rangle\\
\tt A3 = \langle\tt \emptyset,\emptyset,\{a\},\emptyset\rangle
&\tt  B3 = \langle\tt \{a\},\{ (a:b/c) \},\{c\},\emptyset\rangle\\
\tt A4 = \langle\tt \{ d \},\{d:a/a\},\{a\},\emptyset\rangle
&\tt  B4 = \langle\tt \{a\},\emptyset,\{a\},\{a:b/c\}\rangle\\
\tt A5 = \langle\tt \{ a \},\{a:b/c\},\{c\},\emptyset\rangle
&\tt  B5 = \langle\tt \{d\},\{d:c/e\},\{e\},\emptyset\rangle\\
\end{array}
\]
\end{example}

An inconsistent default argument, is inferentially supported by, and inferentially supports, any default argument. 

\begin{example}
Consider 
$\tt A = \langle\tt \{\bot\},\emptyset,\{\bot\},\emptyset\rangle$, 
and $\tt B = \langle\tt \{a\},\{(a:b/b)\},\{b\},\{(b:c/c)\}\rangle$.
So $\tt A$ inferentially supports, directly supports, explicitly supports, and justification supports, $\tt B$,
and $\tt B$ inferentially supports, but not directly suppports, nor explicitly supports, nor justification supports $\tt A$. 
\end{example}

\begin{proposition}
If default argument $A$ directly supports, or explicitly supports, default argument $B$,
then $A$ inferentially supports $B$.
\end{proposition}

\begin{proof}
(1) Assume $A$ directly supports $B$.
So $\ep(B) \in \consequence(A)$.
Since $\ep(B)\in \support(B)$,
$\consequence(A) \cap \support(B) \neq \emptyset$.
(2) Assume $A$ explicitly supports $B$.
So $\cn(\ec(A))\cap\support(B)\neq\emptyset$.
So $\consequence(A) \cap \support(B) \neq \emptyset$.
\end{proof}

Now we turn to notions of attack from one argument on another.
We use the following definition of attacks for a default argument $A$ on default argument $B$ which essentially specifies attack as being an inconsistency between the claim of the attacker and either the support or claim of the attackee. Another kind of attack involves inconsistency between the support of an argument and the justification of the the other argument. In this attacker, the attacker negates the justification of the attackee, and thereby presents a reason to block the use of the default rule.

\begin{definition}
For default arguments $A$ and $B$,
the following are types of attack.
\begin{itemize}[noitemsep]
\item $A$ {\bf support attacks} $B$ 
iff $\consequence(A)\cup\support(B)\vdash\bot$. 
\item $A$ {\bf consequence attacks} $B$ 
iff $\consequence(A)\cup\consequence(B)\vdash\bot$. 
\item $A$ {\bf justification attacks} $B$ 
iff $\consequence(A)\cup\justification(B))\vdash\bot$.
\end{itemize}
\end{definition}

\begin{example}
For the arguments below, 
we have the following relationships:
$\tt A$ support attacks $\tt B$,
$\tt A$ consequence attacks $\tt B$,
$\tt C$ support attacks $\tt D$,
$\tt C$ does not claim attack $\tt D$,
$\tt E$ does not support attack $\tt F$,
$\tt E$ consequence attacks $\tt F$,
$\tt G$ justification attacks $\tt H$,
and
$\tt I$ justification attacks $\tt J$.
\[
\begin{array}{cc}
\tt A = \langle \{a\}, \{a:b/b\}, \{b\}, \{ b:\neg e/ \neg e\}\rangle 
& \tt B = \langle \{d\}, \{d:c \wedge e/e\}, \{e\}, \emptyset \rangle\\
\tt C = \langle \{a\}, \emptyset, \{a\}, \emptyset\rangle 
& \tt D = \langle \emptyset, \{\top:\neg a/\neg a, \neg a:b/b\}, \{b\}, \emptyset \rangle\\
\tt E= \langle \{a\}, \{a:b\wedge c/c\}, \{c\}, \emptyset\rangle 
& \tt F = \langle \{e\}, \{e:f/f\}, \{f\}, \{f:\neg c/\neg c\} \rangle\\
\tt G = \langle \{a\}, \{a:b/b\}, \{b\}, \{ b:\neg c/ \neg c\}\rangle 
& \tt H = \langle \{d\}, \{d:c \wedge e/e\}, \{e\}, \emptyset \rangle\\
\tt I = \langle \{a\}, \{a:b/b\}, \{b\}, \emptyset\rangle 
& \tt J = \langle \{d\}, \{d:\neg b \wedge e/e\}, \{e\}, \emptyset \rangle\\

\end{array}
\]
\end{example}

So a justification attack by an argument negates a justification of default rule used in the attacked argument.
Note, there is no link between support attacks, consequence attacks, or justification attacks.
In other words, it is straightforwards to find examples that are instances of one these relations but that are not instance of either of the other two relations.  

We now consider the following overarching definition of attack by default argument $A$ on default argument $B$ which is a generalization of support attack, claim attack, and justification attack.

\begin{definition}
For default arguments $A$ and $B$,
the {\bf attacks relation} is defined as follows.
\begin{itemize}[noitemsep]
\item $A$ {\bf attacks} $B$ 
iff $\consequence(A)\cup\support(B)\cup\justification(B)\vdash\bot$. 
\end{itemize}
\end{definition}

\begin{table}
\begin{center}
\begin{tabular}{|lr|l|l|}
\hline
Aspect of $A$ that is attacked&  & Attack relation on $A$& Type of attack on $A$\\
\hline
Explicit premise & $\ep(A)$ & Undermine&Support attack\\
Default premise & $\ip(A)$& Undercut&Justification attack\\
\hline
Explicit claim & $\ec(A)$& Rebut&Consequence attack\\
Default claim & $\ic(A)$& Overcut&Justification attack\\
\hline
\end{tabular}
\end{center}
\caption{\label{tab:attacks}Types of focused attack on a default argument $A$ according the aspect of $A$ that is attacked. For instance, undermine is an attack on the explicit premise.}
\end{table}

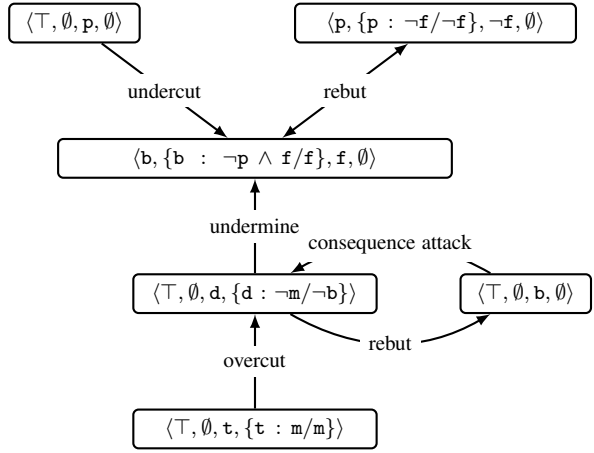
\begin{figure}
\centering
\begin{tikzpicture}
[->,>=latex,thick, scale=0.6,
txtarg/.style={draw,text centered, text width=30mm,
shape=rectangle, rounded corners=2pt,
font=\footnotesize},
]
\node[txtarg] (a0) [text width=15mm] at (0,9) {$\langle\tt \top,\emptyset, p, \emptyset\rangle$};
\node[txtarg] (a1) [text width=50mm] at (4,6) {$\langle\tt b,\{b:\neg p \wedge f/f\}, f, \emptyset\rangle$};
\node[txtarg] (a2) [text width=35mm] at (8,9) {$\langle\tt p,\{p:\neg f/\neg f\}, \neg f, \emptyset\rangle$};
\node[txtarg] (a3) [] at (4,3) {$\langle\tt \top,\emptyset, d, \{ d:\neg m/\neg b\}\rangle$};
\node[txtarg] (a4) [] at (4,0) {$\langle\tt \top,\emptyset, t, \{t:m/m\}\rangle$};
\node[txtarg] (a6) [text width=15mm] at (10,3) {$\langle\tt \top,\emptyset, b, \emptyset\rangle$};
\path	(a0) edge node[fill=white] {\footnotesize undercut} (a1);
\path	(a2) edge[<->] node[fill=white] {\footnotesize rebut} (a1);
\path	(a3) edge node[fill=white] {\footnotesize undermine} (a1);
\path	(a4) edge node[fill=white] {\footnotesize overcut} (a3);
\path	(a6) edge[bend right] node[fill=white] {\footnotesize consequence attack} (a3);
\path	(a3) edge[bend right] node[fill=white] {\footnotesize rebut} (a6);
\end{tikzpicture}
\caption{\label{fig:attackgraph}Instantiated argument map
where $\tt b$ = {\em bird}, 
$\tt p$ = {\em penguin}, 
$\tt d$ = {\em a decoy model that looks like a bird} (i.e. a realistic model of a game bird used by hunters to lure prey to the hunter's position),
$\tt t$ = {\em twitching},
$\tt m$ = {\em moving like a bird},
and $\tt f$ = {\em capable of flying}. 
}
\end{figure}

Given a default argument $B$, we consider four focused ways of attacking a default argument $B$ as described in Table \ref{tab:attacks}. 
We formalize these types of attack as follows,
and we illustrate them in Figure \ref{fig:attackgraph}.

\begin{definition}
For default arguments $A$ and $B$,
the types of {\bf focused attack} are defined as follows.
\begin{itemize}[noitemsep]
\item $A$ {\bf undermines} $B$ 
iff  there exists $\neg\beta \in \consequence(A)$ s.t. $\ep(B) = \beta$.
\item $A$ {\bf rebuts} $B$ 
iff  there exists $\neg\beta \in \consequence(A)$ s.t. $\ec(B) = \beta$.
\item $A$ {\bf undercuts} $B$ 
iff  there exists $\neg\beta \in \consequence(A)$ s.t. there exists $\alpha:\beta/\gamma \in \ip(B)$.
\item $A$ {\bf overcuts} $B$ 
iff  there exists $\neg\beta \in \consequence(A)$ s.t. there exists  $\alpha:\beta/\gamma \in \ic(B)$.
\end{itemize}
\end{definition}

The undermines relation is a special case of the support attack relation, and the rebut relation is a special case of the consequence attack relation. 

We also consider the following special cases of the undermines and rebuts relationships.
For undermines, it is when there is a contradiction between the explicit claim of the attacker and the explicit premise of the attackee;
And for rebuts, it is when there is a contradiction between the explicit claim of the attacker and the explicit claim of the attackee. 

\begin{definition}
For default arguments $A$ and $B$,
some special types of attack are defined as follows.
\begin{itemize}[noitemsep]
\item $A$ {\bf explicitly undermines} $B$ 
iff  there exists $\neg\beta \in \cn(\ec(A))$ s.t. $\beta \in \ep(B)$.
\item $A$ {\bf explicitly rebuts} $B$ 
iff  there exists $\neg\beta \in \cn(\ec(A))$ s.t. $\beta\in\ec(B)$.
\item $A$ {\bf implicitly undermines} $B$ iff $A$ undermines $B$ and $A$ not explicitly undermines $B$.
\item $A$ {\bf implicitly rebuts} $B$ iff $A$ rebuts $B$ and $A$ not explicitly rebuts $B$.
\end{itemize}
\end{definition}

Obviously, explicitly undermines (respectively explicitly rebuts) is a special case of undermines (respectively rebuts). 

\begin{example}
Consider $\tt A = \langle\tt \emptyset,\emptyset, \{a\}, \{ (a:c/c)\}\rangle$
and $\tt B = \langle\tt \emptyset,\emptyset, \{b\}, \{ (b:\neg c/ \neg c)\}\rangle$.
$\tt A$ consequence attacks $\tt B$ but it does not rebut or interdict.
Now consider $\tt C = \langle\tt \emptyset,\emptyset, \{a\wedge b\}, \emptyset\rangle$
and $\tt D = \langle\tt \{\neg a\},\{ (\neg a:\neg c/ \neg c)\}, \{\neg c\}, \emptyset\rangle$.
So $\tt C$ support attacks $\tt D$ but it does not rebut or contradict.
\end{example}

\begin{proposition}
Let $A$ be a default argument where $\ec(A) \vdash \bot$.
For any default argument $B$,
if $\consequence(B)\neq \cn(\{\top\})$, 
then $A$ support attacks $B$,
and if $\support(B)\neq \cn(\{\top\})$, 
then $A$ consequence attacks $B$.
Also, if $\ip(B) \neq \emptyset$, 
then $A$ undercuts $B$,
and if $\cn(\ep(B))\neq \cn(\{\top\})$, 
then $A$ undermines $B$
\end{proposition}

\begin{proof}
Assume $A$ is a default argument where $\ec(A) \vdash\bot$.
If $\consequence(B)\neq \cn(\{\top\})$, 
then there is a $\neg\beta \in \consequence(A)$ s.t. $\beta\in\support(B)$,
and so $A$ support attacks $B$.
If $\support(B)\neq \cn(\{\top\})$, 
then there is a $\neg\beta \in \consequence(A)$ s.t. $\beta\in\consequence(B)$,
and so $A$ consequence attacks $B$.
If $\ip(B) \neq \emptyset$, 
then there is a $\neg\beta \in \consequence(A)$ s.t. $\beta\in\support(B)$, 
and so $A$ undercuts $B$. 
If $\ep(B)\neq \cn(\{\top\})$, 
then there is a $\beta$ s.t. $\beta\in\ep(B)$
and so there is a $\neg\beta \in \consequence(A)$ s.t. $\beta\in\ep(B)$, 
and so $A$ undermines $B$.
\end{proof}

\begin{figure}[t]
\begin{center}
\begin{tikzpicture}
[->,>=latex,thick,
mynode/.style={draw,text centered,rounded corners=3pt, shape=rectangle,fill=blue!10,text width=15mm,font=\footnotesize}, 
]
\node[mynode] (z) [] at (5,7.5) {Attack};
\node[mynode] (a) [] at (1,6) {Support attack};
\node[mynode] (c) [] at (1,4.75) {Undermine};
\node[mynode] (d) [] at (0,3.5) {Explicitly undermine};
\node[mynode] (e) [] at (2,3.5) {Implicitly undermine};
\path	(c) edge node[] {} (a);
\path	(d) edge node[] {} (c);
\path	(e) edge node[] {} (c);
\node[mynode] (a2) [] at (5,6) {Consequence attack};
\node[mynode] (c2) [] at (5,4.75) {Rebut};
\node[mynode] (d2) [] at (4,3.5) {Explicitly rebut};
\node[mynode] (e2) [] at (6,3.5) {Implicitly rebut};
\path	(c2) edge node[] {} (a2);
\path	(d2) edge node[] {} (c2);
\path	(e2) edge node[] {} (c2);
\node[mynode] (a3) [] at (9,6) {Justification attack};
\node[mynode] (b3) [] at (8,4.75) {Undercut};
\node[mynode] (c3) [] at (10,4.75) {Overcut};
\path	(b3) edge node[] {} (a3);
\path	(c3) edge node[] {} (a3);
\path	(a) edge node[] {} (z);
\path	(a2) edge node[] {} (z);
\path	(a3) edge node[] {} (z);
\end{tikzpicture}
\end{center}
\caption{Relationships between different types of attack relation. The source of an arrow is a special case of the target of the arrow.}
\label{ref:attackrelations}
\end{figure}
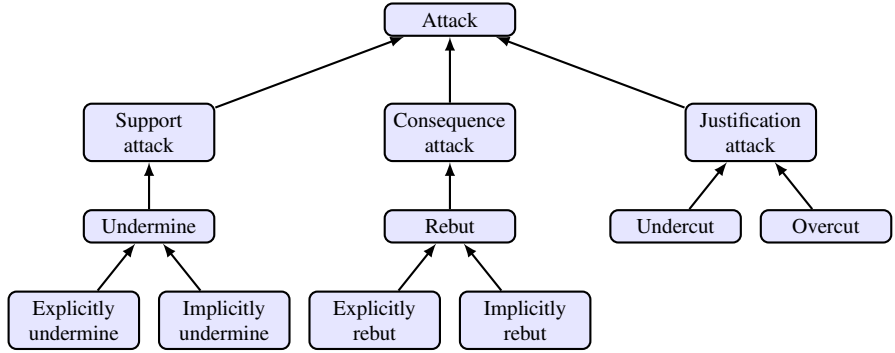

There are some correspondences between different notions of attack between default arguments as we capture in the following results. 
We summarise these results in Figure \ref{ref:attackrelations}. 

\begin{proposition}
For $A$ and $B$ default arguments, the following hold:
(1) If undermines $B$, 
then  $A$ support attacks $B$; 
(2) If $A$ rebuts  $B$, 
then  $A$ consequence attacks $B$;
And 
(3) if $A$ undercuts $B$ or $A$ overcuts  $B$, 
then $A$ justification attacks $B$. 
\end{proposition}

\begin{proof}
(1) If $A$ undermines $B$,  
then there exists $\neg\beta \in \consequence(A)$ s.t. $\beta \in \ep(B)$.
So, $\consequence(A)\cup\support(B)\vdash\bot$, 
and hence $A$ support attacks $B$.
(2) If $A$ rebuts $B$,  
then there exists $\neg\beta \in \consequence(A)$ s.t. $\beta\in\ec(B)$.
So, $\consequence(A)\cup\consequence(B)\vdash\bot$,
and hence $A$ consequence attacks $B$.
(3) If $A$ undercuts $B$, 
then there exists $\neg\beta \in \consequence(A)$ s.t. there exists $\gamma:\beta/\delta \in \ip(B)$,
and if $A$ overcuts $B$, 
then there exists $\neg\beta \in \consequence(A)$ s.t. there exists  $\gamma:\beta/\delta \in \ic(B)$;
So if $A$ undercuts or overcuts  $B$,
there is a $\neg\beta \in \consequence(A)$ s.t. $\beta \in \justification(B)$.
Therefore,  $\consequence(A)\cup\justification(B))\vdash\bot$.
Hence, $A$ justification attacks $B$. 
\end{proof}

\begin{proposition}
For default arguments $A$ and $B$, suppose each rule in $\ic(B)$ is normal: 
(1) If $A$ undercuts $B$, 
then $A$ support attacks $B$;
And (2) if $A$ overcuts $B$, 
then $A$ consequence attacks $B$.
\end{proposition}

\begin{proof}
Assume every rule in  $\ic(B)$ is normal.
So every rule is of the form $\alpha:\beta/\beta$ in $\ic(B)$.
(1) Assume $A$ undercuts $B$.
So there exists $\neg\beta \in \consequence(A)$ s.t. there exists $\alpha:\beta/\beta \in \ip(B)$.
So there exists $\neg\beta \in \consequence(A)$ s.t. $\beta\in\support(B)$.
So $A$ support attacks $B$.
(2) Assume $A$ overcuts $B$.
So there exists $\neg\beta \in \consequence(A)$ s.t. there exists $\alpha:\beta/\beta \in \ic(B)$.
So there exists $\neg\beta \in \consequence(A)$ s.t. $\beta\in\consequence(B)$.
So $A$ consequence attacks $B$.
\end{proof}

\begin{figure}
\centering
\begin{subfigure}{\textwidth}
\centering
\begin{tikzpicture}
[->,>=latex,thick, scale=0.6,
txtarg/.style={draw,text centered, text width=30mm,
shape=rectangle, rounded corners=2pt,
fill=gray!20,font=\footnotesize}]
\node[txtarg] (a) [text width=40mm] at (3,0) {{\rm This sentence is false.}};
\draw (a) .. controls (0,1.5) and (6,1.5) .. (a) node[above,pos=0.5]{$-$};
\end{tikzpicture}
\caption{\label{ex:selfcycle:attack}
Argument graph with a single self-attacking arc.
An example of an instantiation is $\langle \tt \emptyset, \{ \top:a/a \}, \{b\} , \{ b:\neg a/\neg a \} \rangle$
where the atom $\tt a$ denotes ``this sentence is true"
and the atom $\tt b$ denotes ``this sentence is false".
The attack is justification attack. 
}
\end{subfigure}
\begin{subfigure}{\textwidth}
\centering
\begin{tikzpicture}
[->,>=latex,thick, scale=0.6,
txtarg/.style={draw,text centered, text width=30mm,
shape=rectangle, rounded corners=2pt,
fill=gray!20,font=\footnotesize}]
\node[txtarg] (a) [text width=40mm] at (3,0) {{\rm This sentence is true.}};
\draw (a) .. controls (0,1.5) and (6,1.5) .. (a) node[above,pos=0.5]{$+$};
\end{tikzpicture}
\caption{\label{ex:selfcycle:support}
Argument graph with a single self-supporting arc.
An example of an instantiation is  $\langle \tt \emptyset , \{ \top:a/a \}, \{a\} , \emptyset \rangle$
where the atom $\tt a$ denotes ``this sentence is true".
The support is justification support. 
}
\end{subfigure}
\caption{\label{fig:selfcycle}Examples of argument graphs with self-cycles.}
\end{figure}

The expressivity and structure of default arguments gives a wider range of attacks than considered for other proposals for structured argumentation. In particular, the notion of implicit claim rules gives us implicit claims.
It also allows us to consider interesting cases of self-cycles such as in Figure \ref{fig:selfcycle}.

An argument $A$ can both attack and support another argument $B$ since the claim of $A$ can infer a premise or claim of $B$ and also contradict a premise or claim of $B$.

\begin{example}
For arguments $\tt A = \langle \{a\}, \{a:b/c\wedge \neg d\}, \{c\wedge \neg d\}, \emptyset   \rangle$
and $\tt B = \langle \{c\}, \{c:d\wedge e/ d \wedge e\}, \{e\}, \emptyset   \rangle$.
$A$ supports $B$ because $\support(A) \cap \consequence(B) \neq \emptyset$,
and $A$ attacks $B$ because $\tt \neg d \in \consequence(A)$ and $\tt d \in \consequence(B)$.
\end{example}

By using default arguments, we capture a range of ways of defining how one argument attacks another as summarized in Table \ref{tab:attacks}. In particular, we capture attack on the explicit premises (undermines), on the explicit claim (rebuts), on the default derivation of the explicit claim from the premises (undercuts), and on the default derivation of the implicit claim from the explicit claim (overcuts). Furthermore, we capture attack on the inferred premises (interdict) and on the inferred claim (contradict).

\section{Bridging framework}

Now we can address the need to bridge argument maps and logic-based argumentation by producing instantiated argument maps.
The first part of this bridging is the logical translation of premises and claims into logical formulae using a translation function (as discussed in Section \ref{section:translation}). 
The second part of the briding is the instantiation by the logical argument assignment function defined below.
This simply assigns a default argument to each node in the argument map.

\begin{definition}
Let $M = (N,P,C,L)$ be an argument map
and let ${\cal A}$ be a set of default arguments.
A {\bf logical argument assignment} for $M$
is a function $I: N \rightarrow {\cal A}$.
\end{definition}

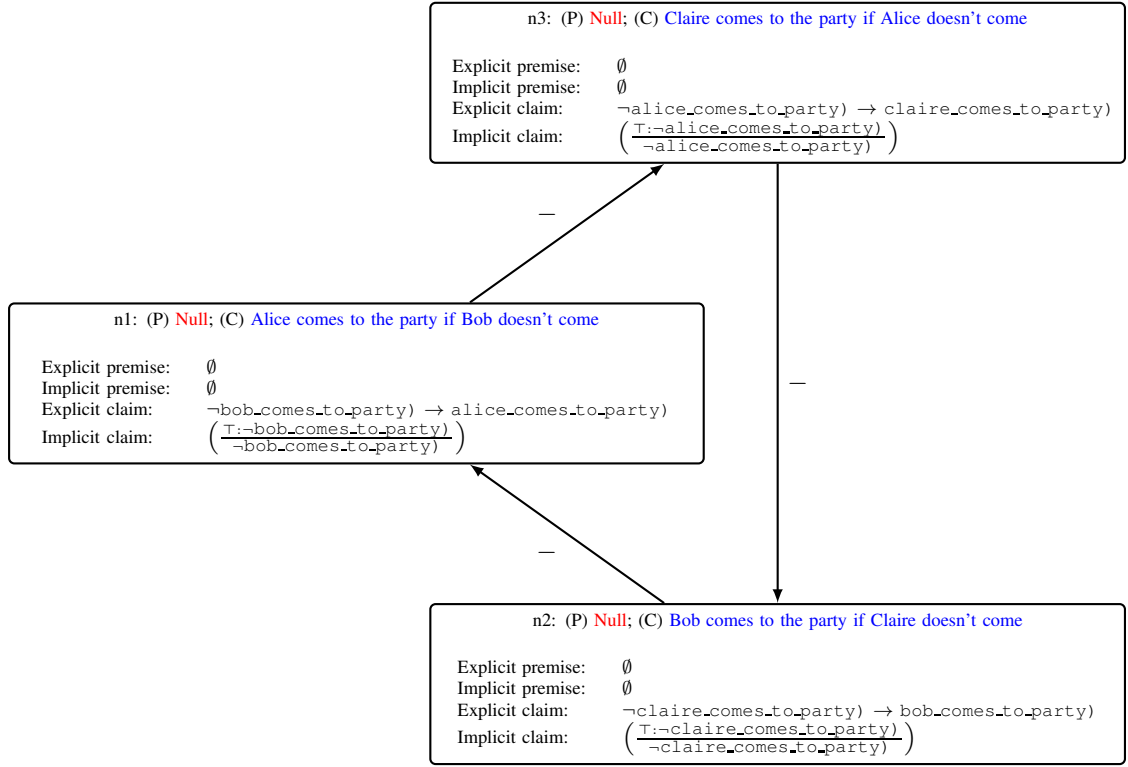
\begin{figure}[t]
\begin{center}
\begin{tikzpicture}[->,>=latex,thick, scale=0.8,
txtarg/.style={draw,text centered, text width=35mm,
shape=rectangle, rounded corners=2pt,
font=\scriptsize}]
\node (a1)[txtarg] [text width=90mm] at (1,5) {n1: (P) \myred{Null}; (C) \myblue{Alice comes to the party if Bob doesn't come}
\[
\begin{array}{ll}
\mbox{Explicit premise: } & \emptyset\\
\mbox{Implicit premise: } & \emptyset\\
\mbox{Explicit claim: } & \neg\mbox{\tt bob\_comes\_to\_party)} \rightarrow \mbox{\tt alice\_comes\_to\_party)} \\
\mbox{Implicit claim: } &\left( \frac{\top:\neg\mbox{\tt bob\_comes\_to\_party)} }{\neg\mbox{\tt bob\_comes\_to\_party)} }\right) \\
\end{array}
\]
};
\node[txtarg] (a2) [text width=90mm] at (8,0) {n2: (P) \myred{Null}; (C) \myblue{Bob comes to the party if Claire doesn't come}
\[
\begin{array}{ll}
\mbox{Explicit premise: }&\emptyset\\
\mbox{Implicit premise: }&\emptyset\\
\mbox{Explicit claim: } & \neg\mbox{\tt claire\_comes\_to\_party)} \rightarrow \mbox{\tt bob\_comes\_to\_party)} \\
\mbox{Implicit claim: } &\left( \frac{\top:\neg\mbox{\tt claire\_comes\_to\_party)} }{\neg\mbox{\tt claire\_comes\_to\_party)} }\right)\\
\end{array}
\]
};
\node[txtarg] (a3) [text width=90mm] at (8,10) {n3: (P) \myred{Null}; (C) \myblue{Claire comes to the party if Alice doesn't come}
\[
\begin{array}{ll}
\mbox{Explicit premise: }&\emptyset\\
\mbox{Implicit premise: }&\emptyset\\
\mbox{Explicit claim: } & \neg\mbox{\tt alice\_comes\_to\_party)} \rightarrow \mbox{\tt claire\_comes\_to\_party)} \\
\mbox{Implicit claim: } &\left( \frac{\top:\neg\mbox{\tt alice\_comes\_to\_party)} }{\neg\mbox{\tt alice\_comes\_to\_party)} }\right)\\
\end{array}
\]
};
\path	(a2) edge node[below left] {$-$} (a1);
\path	(a3) edge node[right] {$-$} (a2);
\path	(a1) edge node[above left] {$-$} (a3);
\end{tikzpicture}
\end{center}
\caption{\label{fig:party}An example of an instantiated argument map involving a three-cycle. 
}
\end{figure}

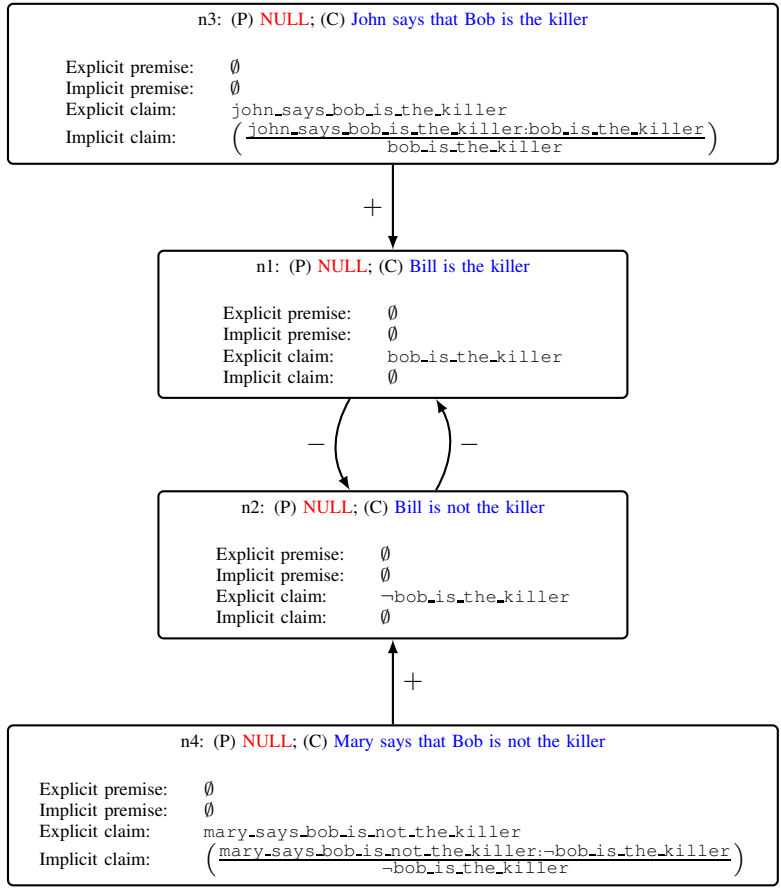
\begin{figure}[t]
\begin{center}
\begin{tikzpicture}[->,>=latex,thick, scale=0.8,
txtarg/.style={draw,text centered, text width=35mm,
shape=rectangle, rounded corners=2pt,
font=\scriptsize}]
\node (a1)[txtarg] [text width=60mm] at (0,8) {n1: (P) \myred{NULL}; (C) \myblue{Bill is the killer}
\[
\begin{array}{ll}
\mbox{Explicit premise: } & \emptyset\\
\mbox{Implicit premise: } & \emptyset\\
\mbox{Explicit claim: } &\mbox{\tt bob\_is\_the\_killer}  \\
\mbox{Implicit claim: } &\emptyset\\
\end{array}
\]
};
\node[txtarg] (a2) [text width=60mm] at (0,4) {n2: (P) \myred{NULL}; (C) \myblue{Bill is not the killer}
\[
\begin{array}{ll}
\mbox{Explicit premise: }&\emptyset\\
\mbox{Implicit premise: }&\emptyset\\
\mbox{Explicit claim: } & \neg \mbox{\tt bob\_is\_the\_killer} \\
\mbox{Implicit claim: } &\emptyset\\
\end{array}
\]
};
\node[txtarg] (a3) [text width=100mm] at (0,12) {n3: (P) \myred{NULL}; (C) \myblue{John says that Bob is the killer}
\[
\begin{array}{ll}
\mbox{Explicit premise: }&\emptyset\\
\mbox{Implicit premise: }&\emptyset\\
\mbox{Explicit claim: } & \mbox{\tt john\_says\_bob\_is\_the\_killer} \\
\mbox{Implicit claim: } & \left( \frac{\mbox{\tt john\_says\_bob\_is\_the\_killer}: \mbox{\tt bob\_is\_the\_killer}}{\mbox{\tt bob\_is\_the\_killer} }\right)  \\
\end{array}
\]
};
\node[txtarg] (a4) [text width=100mm] at (0,0) {n4: (P) \myred{NULL}; (C) \myblue{Mary says that Bob is not the killer}
\[
\begin{array}{ll}
\mbox{Explicit premise: }&\emptyset\\
\mbox{Implicit premise: }&\emptyset\\
\mbox{Explicit claim: } & \mbox{\tt mary\_says\_bob\_is\_not\_the\_killer} \\
\mbox{Implicit claim: } & \left( \frac{\mbox{\tt mary\_says\_bob\_is\_not\_the\_killer}:\neg \mbox{\tt bob\_is\_the\_killer} }{\neg\mbox{\tt bob\_is\_the\_killer} }\right)  \\
\end{array}
\]
};
\path	(a1) edge[bend right] node[left] {$-$} (a2);
\path	(a2) edge[bend right] node[right] {$-$} (a1);
\path	(a3) edge node[left] {$+$} (a1);
\path	(a4) edge node[right] {$+$} (a2);
\end{tikzpicture}
\end{center}
\caption{\label{fig:murder}An example of an instantiated argument map adapted from \cite{Prakken2023}.
}
\end{figure}

We give examples of logical argument assignments 
in Examples \ref{ex:atomic:1} to \ref{ex:fo:1}.
We give further examples in Figure \ref{fig:party} and Figure \ref{fig:murder}. 

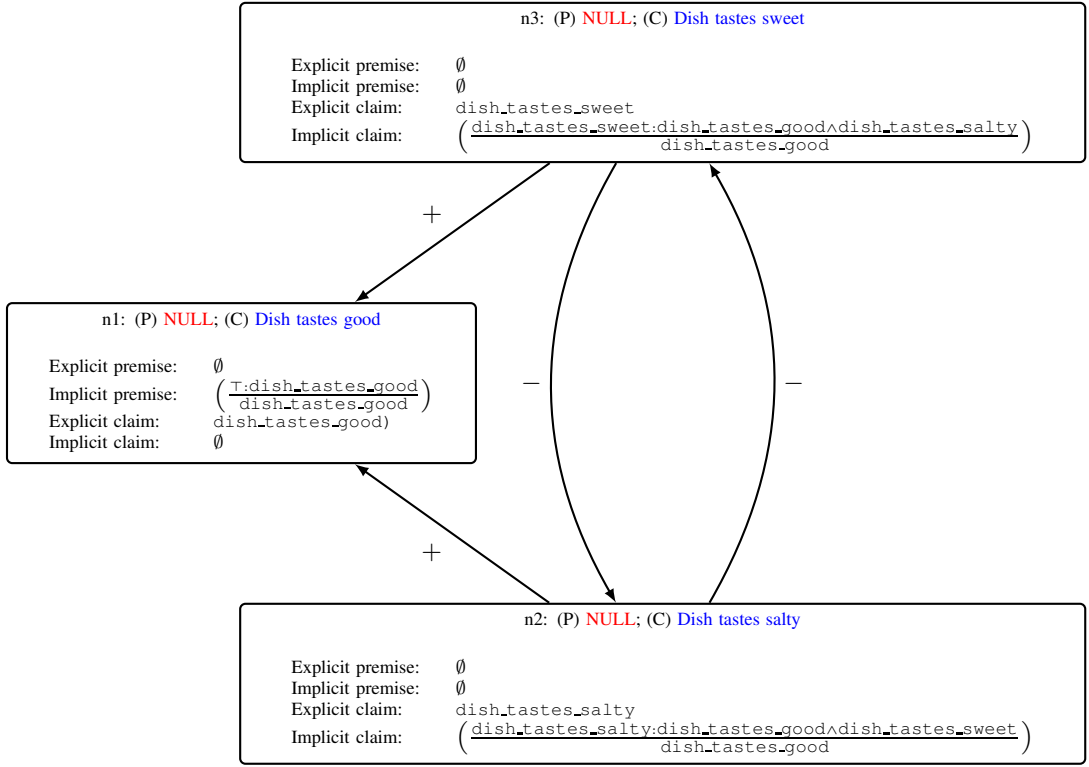
\begin{figure}[t]
\begin{center}
\begin{tikzpicture}[->,>=latex,thick, scale=0.8,
txtarg/.style={draw,text centered, text width=35mm,
shape=rectangle, rounded corners=2pt,
font=\scriptsize}]
\node (a1)[txtarg] [text width=60mm] at (1,5) {n1: (P) \myred{NULL}; (C) \myblue{Dish tastes good}
\[
\begin{array}{ll}
\mbox{Explicit premise: } & \emptyset\\
\mbox{Implicit premise: } & \left( \frac{\top:\mbox{\tt dish\_tastes\_good}}{\mbox{\tt dish\_tastes\_good}}\right)\\
\mbox{Explicit claim: } & \mbox{\tt dish\_tastes\_good)} \\
\mbox{Implicit claim: } &\emptyset\\
\end{array}
\]
};
\node[txtarg] (a2) [text width=110mm] at (8,0) {n2: (P) \myred{NULL}; (C) \myblue{Dish tastes salty}
\[
\begin{array}{ll}
\mbox{Explicit premise: }&\emptyset\\
\mbox{Implicit premise: }&\emptyset\\
\mbox{Explicit claim: } & \mbox{\tt dish\_tastes\_salty} \\
\mbox{Implicit claim: } &\left( \frac{\mbox{\tt dish\_tastes\_salty}:
                \mbox{\tt dish\_tastes\_good} \wedge \mbox{\tt dish\_tastes\_sweet}}
{\mbox{\tt dish\_tastes\_good}}\right)\\
\end{array}
\]
};
\node[txtarg] (a3) [text width=110mm] at (8,10) {n3: (P) \myred{NULL}; (C) \myblue{Dish tastes sweet}
\[
\begin{array}{ll}
\mbox{Explicit premise: }&\emptyset\\
\mbox{Implicit premise: }&\emptyset\\
\mbox{Explicit claim: } & \mbox{\tt dish\_tastes\_sweet} \\
\mbox{Implicit claim: } &\left( \frac{\mbox{\tt dish\_tastes\_sweet}:
                \mbox{\tt dish\_tastes\_good} \wedge \mbox{\tt dish\_tastes\_salty}}
{\mbox{\tt dish\_tastes\_good}}\right)\\
\end{array}
\]
};
\path	(a2) edge node[below left] {$+$} (a1);
\path	(a3) edge node[above left] {$+$} (a1);
\path	(a2) edge[bend right] node[right] {$-$} (a3);
\path	(a3) edge[bend right] node[left] {$-$} (a2);
\end{tikzpicture}
\end{center}
\caption{\label{fig:dish}An example of an instantiated argument map concerning a dish tasting good. 
}
\end{figure}

\begin{example}
\label{ex:atomic:1}
Consider the argument map $M = (N,A,P,C,L)$ (in Figure \ref{fig:dish}) 
with $N = \{n_1,n_2,n_3\}$ 
where $P(n_1) = NULL$, $C(n_1) = \mbox{\em Dish tastes good}$; 
$P(n_2) = NULL$, $C(n_2) = \mbox{\em Dish tastes salty}$;
$P(n_3) = NULL$, $C(n_3) = \mbox{\em Dish tastes sweet}$; 
such that $L(n_2,n_1) = L(n_3,n_1) = +$ and $L(n_3,n_2) = L(n_2,n_3) = -$.
Moreover, suppose an atomic logical translation function $Ta$ such that 
$T(P(n_1)) = \emptyset$, $T(C(n_1)) = \tt Dish\_tastes\_good$,
$T(P(n_2)) = \emptyset$, $T(C(n_2)) = \tt Dish\_tastes\_salty$,
$T(P(n_3)) = \emptyset$, $T(C(n_3)) = \tt Dish\_tastes\_sweet$.
\end{example}

When instantiating an argument map, we would aim to instantiate the nodes so as to respect the labels on the arc. In other words, if an arc is labelled $+$, then the source argument should be a supporter of the target argument, and if an arc is labelled $-$, then the source argument should be an attacker of the target argument.
For example, for Figure \ref{fig:dish},
$\tt n2$ and $\tt n3$ are supporters of $\tt n1$, 
and $\tt n2$ attacks $\tt n3$ and vice versa.

We give the following definition of a specific instantiation function to illustrate how we can systematically specify how an argument map should be undertaken.
In the definition we propose to instantiate defaults arguments only by relations targeting explicit
premises (attacks = explicitly undermines and supports = directly supports).

\begin{definition}
\label{def:premiseatomiclogicalargument}
Let $M = (N,P,C,L)$ be an argument map, let ${\cal A}$ be a set of default arguments, 
For each $n_i \in N$, the atomic logical translation function $T$ 
is such that $T(P(n_i)) = a_i$, $T(C(n_i)) = b_i$.
The {\bf premise atomic logical argument assignment} for $M$ is the smallest function $I: N \rightarrow {\cal A}$, such that
for all $n_i,n_j\in N$, 
\begin{itemize}
    \item $\ip(I(n_i)) = \{(a_i:b_i/b_i)\}$;
    \item if $n_i$ attacks $n_j$ (i.e. $L(n_i, n_j) = -$), then $(b_i: \neg a_j/\neg a_j) \in \ic(I(n_i))$;
    \item if $n_j$ supports $n_j$ (i.e. $L(n_i, n_j) = +$), then $(b_i: a_j/ a_j) \in \ic(I(n_i))$.
\end{itemize}
\end{definition}

\begin{example}
\label{ex:atomicassignment}
Let $M = (N,P,C,L)$ be an argument map where $N = \{n_1,n_2,n_3\}$ s.t. $L(n_2,n_1) = +$ and $L(n_3,n_1) = -$
and where the atomic logical translation function $T$ is 
such that $T(P(n_1)) = \{\tt a_1\}$, $T(C(n_1)) = \{\tt b_1\}$,
$T(P(n_2)) = \{\tt a_2\}$, $T(C(n_2)) = \{\tt b_2\}$,
$T(P(n_3)) = \{\tt a_3\}$, $T(C(n_3)) = \{\tt b_3\}$.
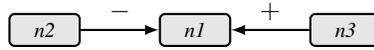
\begin{center}
\begin{tikzpicture}
[->,>=latex,thick, 
txtarg/.style={draw,text centered, text width=7mm,
shape=rectangle, rounded corners=2pt,
fill=gray!20,font=\footnotesize},
]
\node[txtarg] (a1) [] at (2,0) {n1};
\node[txtarg] (a2) [] at (0,0) {n2};
\node[txtarg] (a3) [] at (4,0) {n3};
\path	(a2) edge node[above] {$-$} (a1);
\path	(a3) edge node[above] {$+$} (a1);
\end{tikzpicture}
\end{center}
Given this argument map, a premise atomic logical argument assignment $I$ give the following assignment to the arguments.
\[
\begin{array}{l}
I(n_1) = \langle\tt \{a_1\},\{(a_1:b_1/b_1)\}, \{b_1\}, \emptyset\rangle.\\ 
I(n_2) = \langle\tt \{a_2\},\{(a_2:b_2/b_2)\}, \{b_2\}, \{(b_2:\neg a_1/\neg a_1)\}\rangle.\\
I(n_3) = \langle\tt \{a_3\},\{(a_3:b_3/b_3)\}, \{b_3\}, \{(b_3:a_1/a_1)\}\rangle.\\
\end{array}
\]
\end{example}

We have many choices for instantiation functions based on how we construct the implicit and explicit aspects of each argument. 
When defining a systematic way of instantiation function, we need to decide what aspects of each argument are reflect by a formula in the explicit premise or explicit claim, and what aspects are reflected by a default rule in the implicit premise or claim. Furthermore, we need to consider how arguments can be supported or attacked (i.e. the choice of definition for support or attack), and this will also have implications for how we use the justifications for the default rules. 

Some general options for instantiation of default argument include:
An {\bf atomic instantiation} which assigns an argument where $\alpha$ and $\beta$ are atoms and $\Phi$ and $\Psi$
are sets of default rules of the form $(\gamma:\delta/\delta)$ where $\gamma$ and $\delta$ are atoms (Example \ref{ex:atomic:1});
A {\bf propositional instantiation} which assigns an argument where $\alpha$ and $\beta$ are propositional formula and $\Phi$ and $\Psi$ are sets of default rules of the form $(\gamma:\delta/\delta)$ where $\gamma$ and $\delta$ are propositional formula (Example \ref{ex:propositional:1});
And a {\bf first-order instantiation} which assigns an argument where $\alpha$ and $\beta$ are first-order formula and $\Phi$ and $\Psi$ are sets of default rules of the form $(\gamma:\delta/\delta)$ where $\gamma$ and $\delta$ are first-order formula (Example \ref{ex:fo:1}).

\begin{example}
\label{ex:propositional:1}
We continue with the argument map in Example \ref{ex:atomicassignment}. 
The following is a propositional instantiation. 
\[
\begin{array}{l}
I(n_1) = 
\langle \tt \{c\},\{(c: d \wedge (e \rightarrow f)/ e \rightarrow f)\}, \{e \rightarrow f\}, \{(\top : e/e)\}\rangle.\\
I(n_2) = 
\langle \tt \{a \vee b\}, \{(a \vee b \vee c: \neg d / \neg d) \}, \{\neg d\}, \emptyset\rangle.\\
I(n_3) = 
\langle \tt \{g, g \rightarrow h\}, \emptyset, \{h,h\rightarrow c\}, \emptyset\rangle.\\
\end{array}
\]
\end{example}

\begin{example}
\label{ex:fo:1}
We continue with the argument map in Example \ref{ex:atomicassignment}. 
The following is a first-order instantiation. 
\[
\begin{array}{l}
I(n_1) = 
\langle \tt \{\forall x,y \; m(x,y)\},\emptyset, \{\forall x,y \; m(x,y)\}, \emptyset\rangle.\\
I(n_2) = 
\langle \tt \{\exists x,y \; \neg m(x,y)\},\emptyset,  \{\exists x,y \; \neg m(x,y)\}, \emptyset\rangle.\\
I(n_3) = 
\langle \tt \emptyset,\emptyset, \{ \forall x,y \; m(x,y)\}, \emptyset\rangle.\\
\end{array}
\]
\end{example}

There is a wide variety of instantiation functions that we may consider. We can restrict the codomain by the choice of expressivity of language (e.g. the set of universally quantified formulae without function symbols, or the set of first-order predicate formulae composed from binary relations --- and thereby draw on a knowledge graph). Furthermore, we can restrict the non-logical symbols and so restrict what atoms or predicates, constants, etc we use.

The advantage of the instantiation based on Definition \ref{def:premiseatomiclogicalargument} is that we have a relatively small number of default rules to acquire and use.
If cardinality of ${\cal A}$ is $k$, then the cardinality of ${\cal R}$ is $2 \times k \times k$.
This is because each default rule is of the form $(a_i:b_j/b_j)$ or $(a_i:\neg b_j/\neg b_j)$.
So there $k$ choices for $a_i$ and $k$ choices for $b_j$, and then for each choice of $a_i$ and $b_j$, there are two default rules; the first with $b_j$ and the second with $\neg b_j$ as the justification and consequence. 
This means that if we have a set of examples of cardinality for testing each default rule, we determine the probability of each default being correct in polynomial time. 

To restrict instantiation, we can consider a specific knowledge-base $(\Pi,\Gamma)$
where $\Pi$ is a set of default rules and $\Gamma$ is a set of classical formulae.
An argument $A$ is based on a knowledge-base  $(\Pi,\Gamma)$
iff $\ep(A)\subseteq \Gamma$ 
and $\ec(A)\subseteq \Gamma$
and $\ip(A)\subseteq\Pi$ 
and $\ic(A)\subseteq \Pi$.
Let $\args(\Pi,\Gamma)$ be the set of arguments based on $(\Pi,\Gamma)$.

In general, if $|\args(\Pi,\Gamma)| = k$, and $|\nodes(M)| = n$,
then there are $n^k$ instantiation functions.
This is because for each node there are $k$ choices of argument to instantiate the node, and there are node nodes, hence $n^k$ instantiations. 
In practice, there will be far fewer instantiation functions which satisfy constraints on arguments such as validity, mininality, coherence, and on relationships between arguments satisfying labels.

A knowledge-base $(\Pi,\Gamma)$ is {\bf normal exhaustive} iff for every $\alpha,\beta\in \Gamma$,
$(\alpha:\beta/\beta) \in \Pi$. 
In the case of atomic instantiations, 
if $\Pi$ is normal exhaustive, 
and $|\Gamma| = x$, 
then $|\Pi| = (x+1)^2$ (because any atom or $\top$ can be $\alpha$ or $\beta$).

We now have a formal framework that specifies the input-output relationship for bridging argument maps and logic-based argumentation: The input is the argument map (from argument mining), and the output is the instantiated argument map. 
Furthermore, our new framework provides a logical representation of both the implicit and explicit aspects of the premises and claims.
This can then be used to investigate ambiguity arising from implicitness which results in choices for an instantiation of an argument map.
By analysing the range and diversity of possible instantiations for an argument map, we may be able to measure the rebustness of any part of an instantiation.
We may also consider constraints on instantiations that capture desirable behavour 
(such as been proposed for instantiating bipolar argument graphs with deductive arguments \cite{Hunter2023}.
The framework may also be viewed as a starting point for deciding what is an semantically appropriate instantiation 
by using some kind of semantic analysis of the explicit premises, explicit claims and the relations.

\section{Literature review}

In this section, we consider our proposal for instantiated argument maps with developments in abstract bipolar argumentation, structured argumentation including approaches using default logic, and formalisms for handling of enthymemes.

\subsection{Bipolar argumentation}

Bipolar argumentation is a generalization of abstract argumentation that incorporates a support relation in addition to the attack relation \cite{CayrolLS05,CayrolLS05b,Amgoud2008,CayrolLS13}. 
A bipolar argument graph is a directed graph where each node  denotes an argument, and each arc denotes the influence of one argument on another.
The label denotes the type of influence with options including positive (supporting) and negative (attacking).
In order to determine the acceptable arguments from a biploar argument graph, dialectical semantics can be generalized to handle both support and attack relations \cite{CayrolLS13}. Alternatively, the approach of abstract dialectical frameworks \cite{Brewka2010,Brewka2014}, gradual semantics \cite{Amgoud2013,AmgoudBNDV17,amgoud2021general, Bonzon16,CayrolLS05b,LeiteMartins11,Rago16,Costa-Pereira:2011,BRTAB15,Potyka18,Pu2014,PuLZL15,Baroni2019}, or epistemic graphs \cite{Hunter2020} can be used.

An issue with bipolar argument graphs is that each node is an abstract argument, and so the meaning of the argument is unspecified. To address this issue, we have investigated how nodes can be instantiated with default arguments. By doing this, we can systematically investigate the nature of the contents (premises and claim) and their interplay with the structure of the graph.  
But this then raises questions about what kinds of instantiations are appropriate and what they  mean. It also raises questions about how we can compare instantiated bipolar argument graphs to show, for instance, that two instantiations are equivalent, and how we can investigate the interplay of the premises of an argument and the claims of the arguments that support or attack it.

\subsection{Structured argumentation}

Our proposal goes beyond existing proposals for structured argumentation. In Section \ref{section:defaultarguments}, we have shown how we can capture classical logic deductive arguments as default logic. It is then straightforward to capture the range of attack relationships \cite{BH01,BH08,Gorogiannis2011} and support relationships \cite{Hunter2023} in our new framework. Whilst default knowledge can be captured in the deductive argumentation approach using normality atoms \cite{BH14}, our new proposal provides a more systematic way of representing these as justifications. 

The two main approaches to structured argumentation 
are assumption-based argumentation (ABA) \cite{Toni2014}, and ASPIC+ \cite{Modgil2014}. 
In ABA, assumptions are atoms that can be used in the condition of proof rules in a way that replicates justifications in default rules. 
So a proof rule can be attacked when the contrary of an assumption holds.
Furthermore, assumption-based argumentation has been shown to capture default lotic \cite{Bondarenko1997}. In other words, any default theory can be equivalently represented by an ABA knowledge-base. 
Similarly to ASPIC+, proof rules are labelled, and these labels can be used at atoms in the formulae of the object language. Furthermore, an argument using the proof rule is attacked when the contrary of its label holds.
In contrast to ASPIC+ and ABA, our proposal in this paper clearly demarks the implicit premises and implicit claims in a strutured argument representation of an enthymeme, and it further goes beyond ASPIC+ and ABA by providing a framework of definitions for support and attack relations between default arguments.

There are other proposals in the computational argumentation literature for using default logic such as \cite{Prakken1993,Santos2008,Hunter2018}. But our proposal goes beyond these in that we provide a comprehensive coverage of differnt types of argument and of different types of support and attack relationships. 
In the proposal by Prakken \cite{Prakken1993}, a default argument is based on a ground default theory with a unique extension where the premises are the ground default theory and the claim is an element in the extension.
An argument $A$ is a counterargument to an argument $B$ if the claim of $A$ contradicts $B$. These notions of argument and counterargument are special cases in our more general framework. 
In the proposal by Santos {\em et al.} \cite{Santos2008} for using default logic in argumentation, 
the premises of an argument is a minimal default theory $(D,W)$, with a unique and consistent extension, where the claim is in the extension of the default theory. They also introduce the notion of a justificative argument which is an argument based on a default theory $(D,W)$ where the justifications of each default in $D$ is implied by $W$. A justificative argument can be viewed as an argument that does not rely on ``unknown" information (i.e. the argument can be written into deductive argumentation where the justification can be equally represented as a precondition). Furthermore, for an argument with premises $(D,W)$, an undercut is defined as an argument with a claim that negates a some of the formulae in $W$ and and/or some of the justifications in the rules in $D$. 
The proposals by Prakken and by Santos {\em et al.} focus on specific kinds of arguments, and do not consider the more general notion of arguments that we present in this paper. Furthermore, they do not give consideration to handling implicit premises and implicit claims, and so do not consider issues of modelling enthymemes.

\subsection{Handling enthymemes}

Most proposals for handling enthymemes in the compuptational argumentation literature either focus on the coding/decoding of enthymemes via abduction \cite{Hunter2007,Black2012,Hosseini2014},
and how this can be undertaken within a dialogue \cite{Black2008,Dupin2011,Dupin2011b,Xydis2020,Panisson2022,Leiva2023}.
For instance, dialogues can be shown to shorter when using enthymemes \cite{Panisson2022}.
However, these proposals do not provide a systematic framework for translating argument maps into logic, and concomitantly, addressing the problem of identifying the missing premises and/or claim, and discerning the relationships between them. 

Another important aspect of dealing with enthymemes is the uncertainty that arises from decoding. 
When an audience is listening to participants in a discussion or debate, 
the participants present arguments including enthymemes.
The way these are presented gives some idea to the audience of which are meant to attack which. However, if an argument being attacked is an enthymeme, and the attacking and attacked arguments are from different participants, then there is uncertainty about whether it is indeed a valid attack. Each enthymeme is a representative for an intended argument, but for the audience it may be uncertain which decoding is the intended argument. The audience may have zero or more choices. This means that the audience takes an argument graph as input, and tries to determine the intended argument graph (i.e. the graph obtained by instantiating each node with its intended argument and deleting the arcs that are not valid attacks). This intended argument graph has a structure that would be isomorphic to a spanning subgraph of the original argument graph. 
One way to model this uncertainty is to use the constellations approach to probabilistic argumentation, where a probability distribution over these graphs is used to reflect the uncertainty \cite{Hunter2013}. However, this assumes that the graph contains abstract arguments. Another approach that is also based on abstract argumentation. is to use the approach of incomplete graphs \cite{Mailly2016}. 

A key problem with these proposals for handling enthymemes in the computational argumentation literature 
(such as \cite{Hunter2007,Black2012,Hosseini2014,Xydis2020,Panisson2022}) 
is that they assume an extensive set of common or commonsense knowledge with a preference ordering over the appropriateness of any formula for a specific audience. These are demanding assumptions. It is difficult to acquire commonsense knowledge, and in particular, there are often gaps in the explicit knowledge that connects concepts. 
Relationships between concepts are often known but they are implicit. For instance, different symbols for the same concept are used (e.g. $\qw{dad}$ and $\qw{father}$) or for similar concepts ($\qw{showery}$ and $\qw{rainy}$), though sometimes such correspondences are context dependent (e.g. in the context of a church,  $\qw{dad}$ and $\qw{father}$ often refer to different concepts). 

To address this need for a scalable and robust way of connecting concepts,  distance measures between semantic concepts can be used \cite{HunterAAAI2022}. 
We assume that each semantic concept is represented by a word, and so in propositional logic, each propositional atom is a word  (or compound word), and in predicate logic, each predicate, function, and constant, symbol is a word (or compound word). So with a distance measure, we want the distance between $\qw{dad}$ and $\qw{father}$ to be low, and between $\qw{dad}$ and $\qw{grass}$ to be high. Fortunately, there are now some robust and scalable options for distance measures that we can use based on word embeddings and sentence embeddings. None of these will be totally correct, but the error rate could be tolerable, and the benefits of using them far outweighing this. 

\subsection{Translating text into logic}

Within the NLP community, there has always been interest in technology for translating free text into formal logic. In the past, use of approaches such as phrase-structure grammars, allows for translation of fragments of natural language into logic, but it is only with the  advent of deep leanring and large language models that it appears more general, robust, and scalable, methods can be developed for translating natural language into logic. Preliminary investigations using
deep learning include \cite{Singh2020,Levkovskyi2021} and using LLMs include \cite{Lu2022,Pan2023,Olausson2023,Lalwani2024}. 

Another approach is to use abstract meaning representation (AMR) as a semantic representation of a sentence as a directed graph. The aim of AMR is that the graph reflects the content of a sentence in terms of the actors, and their roles, in the sentence, rather than a complete coverage of the grammar of the sentence \cite{Langkilde1998}. So it is intended that sentences that are grammatically different but the same or similar semantics are represented by the same AMR graph. By drawing on LLMs, some AMR parsers offer a general, robust, and scalable, approach to generating AMR graphs from a wide variety of text (e.g. \cite{Lee2023}). Once, an AMR graph has been obtained from free text, a logical formula can be obtained which can be reasoned with using neurosymbolic approach \cite{Chanin2023,Feng2024}. 

Finally, a recent proposal for analysing enthymemes draws on the semi-formal approach of argument schema \cite{RuizDolz2024}. First, a dataset has been synthetically generated using large language models (GPT3.5 and GPT 4) where the model is prompted to fill out an argument scheme from a selection of 20 schema. Then, fine-tuned Roberta models were developed for classification of free text arguments according the type of argument scheme. Whilst, this proposal does not produce logical arguments, it could be harnessed as part of process of producing logical arguments. This dataset has also been used in developing a reinforcement learning approach for classifying free text arguments according the type of argument scheme using Mistral \cite{Trajano2024}.

\section{Discussion}

In the real-world, arguments are often enthymemes (i.e. arguments with some premises being implicit).
The missing premises have to be abduced by the recipient. These may be obtained from contextual and
background knowledge, and more challengingly, common-sense knowledge.
We need to understand enthymemes if we want to make sense of them or respond to them. 
If we don't understand an enthymeme properly, we can have misunderstandings, 
and we can talk at cross-purposes.

Argument mining find sentences representing premises and claims. But there is a lack of a framework to then identify the underlying logical arguments.
We have addressed this shortcoming by providing the following novel contributions in this paper: 
(1) A form of structured argumentation based on default logic; 
(2) A framework for representing argument maps and representing logical instantiation of them using structured argumentation based ond default logic;
and
(3) Investigation of specific kinds of instantiation based on types of argument, attack and support, and instantiation method. 

Following from our proposal, there are a number of areas of future work.
First, we would like to investigate methods for translation of premises and claims in logic.
A simple option is to use a large language model to classify each type of premise or claim,
and thereby obtain an atomic translation.
A more complex option is to use an AMR parser to translate text into an AMR graph which in turn can be translated into a propositional or first order formula \cite{Chanin2023,Feng2024}.

Second, we would like to investigate how to handle the uncertainty arising in decoding enthymemes.
This includes the uncertainty from translating the premises and claims into logic and the uncertainty in finding the instantiation.
For instance, if we use a text classifier, then we can have a probability distribution over the classifications, and this could be reflected by a distribution over the instantiations.
There is also the related question of finding instantiations that satisfying the labels (an attacker for a negative label and a supporter for a positive label). This may then lead to a constraint satisfaction problem for satisfying the maximum number of labels. 

Third, we would like to investigate how we can acquire and represent common and common-sense knowledge in the form of default logic. 
It would be desirable to render the approach scalable which therefore calls for automated means for acquiring this kind of knowledge.


\bibliography{enthymeme}
\bibliographystyle{alpha}


\end{document}